\documentclass[11pt]{article}
\usepackage[margin=1in]{geometry}
\usepackage{authblk}
\usepackage{hyperref}
\usepackage{graphicx}
\usepackage{amsmath, amssymb}
\usepackage{booktabs}
\usepackage[table,xcdraw]{xcolor}
\usepackage{rotating}
\usepackage[skip=6pt]{caption} 

\usepackage{multirow}
\usepackage{makecell}
\usepackage{float}
\usepackage{svg}
\usepackage{appendix}
\usepackage{soul}
\usepackage{comment}
\usepackage{listings}

\definecolor{niceorange}{HTML}{F86624}
\definecolor{niceblue}{HTML}{072AC8}
\newcommand\uijletalorig[1]{\textbf{Uijl et al}$_\text{orig-#1}$}
\newcommand\uijletalstruct[0]{\textbf{Uijl et al}$_\text{struct}$}
\newcommand\ebmstruct[0]{\textbf{EBM}$_\text{struct}$}
\newcommand\lrtfidf[0]{\textbf{LR-TF-IDF}}
\newcommand\ebmtfidf[0]{\textbf{EBM-TF-IDF}}
\newcommand\robbert[0]{\textbf{RobBERT}}
\newcommand\medrobertanl[0]{\textbf{MedRoBERTa.nl}}
\newcommand\geitje[0]{\textbf{GEITje}}
\newcommand\auggamlr[0]{\textbf{Aug-Linear}$_\text{LR}$}
\newcommand\auggamebm[0]{\textbf{Aug-Linear}$_\text{EBM}$}
\newcommand\auggamlrstruct[0]{\textbf{Aug-Linear}$_\text{LR+struct}$}
\newcommand\auggamebmstruct[0]{\textbf{Aug-Linear}$_\text{EBM+struct}$}

\begin{document}
%TC:ignore

\title{Interpretable phenotyping of Heart Failure patients with Dutch discharge letters}

\author[1]{Vittorio Torri\thanks{Corresponding author: \href{mailto:vittorio.torri@polimi.it}{vittorio.torri@polimi.it}}}
\author[2,3,4,5]{Machteld J. Boonstra}
\author[2,6]{Marielle C. van de Veerdonk}
\author[2,5,7]{Deborah N. Kalkman}
\author[2,4,8,9,10]{Alicia Uijl}
\author[1,11]{Francesca Ieva}
\author[3,7,12]{Ameen Abu-Hanna}
\author[2,8,13]{Folkert W. Asselbergs}
\author[3,14]{Iacer Calixto}

\affil[1]{MOX - Modelling and Scientific Computing Lab, Department of Mathematics, Politecnico di Milano, Milano, Italy}
\affil[2]{Department of Cardiology, Amsterdam University Medical Center, University of Amsterdam, Amsterdam, The Netherlands}
\affil[3]{Department of Medical Informatics, Amsterdam University Medical Center, University of Amsterdam, Amsterdam, The Netherlands}
\affil[4]{Amsterdam Public Health, Digital Health, Personalized Medicine, Amsterdam, The Netherlands}
\affil[5]{Amsterdam Cardiovascular Sciences, Atherosclerosis and Aortic Disease, Cardiomyopathy and Arrhythmia, Amsterdam, The Netherlands}
\affil[6]{Amsterdam Cardiovascular Sciences, Pulmonary Hypertension and Critical Care, Amsterdam, The Netherlands}
\affil[7]{Amsterdam Reproduction \& Development, Amsterdam, The Netherlands}
\affil[8]{Amsterdam Cardiovascular Sciences, Cardiomyopathy and Arrhythmia, Amsterdam, The Netherlands}
\affil[9]{Department of Epidemiology and Data Science, Amsterdam University Medical Center, University of Amsterdam, Amsterdam, The Netherlands}
\affil[10]{Department of Clinical Science and Education, Södersjukhuset, Karolinska Institutet, Stockholm, Sweden}
\affil[11]{HDS - Health Data Science Centre, Human Technopole, Milano, Italy}
\affil[12]{Amsterdam Public Health, Methodology, Aging \& Later Life, Amsterdam, The Netherlands}
\affil[13]{Institute of Health Informatics, University College London, London, United Kingdom}
\affil[14]{Amsterdam Public Health, Methodology, Mental Health, Amsterdam, The Netherlands}

\date{}

\maketitle

\pagebreak
%Abstracts must be able to stand alone and so cannot contain citations to
% the paper's references, equations, etc. An abstract must consist of a single
% paragraph and be concise. Because of online formatting, abstracts must appear
% as plain as possible.
\abstract{
\noindent\textbf{Objective:} Heart failure (HF) patients present with diverse phenotypes affecting treatment and prognosis. This study evaluates models for phenotyping HF patients based on left ventricular ejection fraction (LVEF) classes, using structured and unstructured data, assessing performance and interpretability.\\
\textbf{Materials and Methods:} The study analyzes all HF hospitalizations at both Amsterdam UMC hospitals (AMC and VUmc) from 2015 to 2023 (33,105 hospitalizations, 16,334 patients). Data from AMC were used for model training, and from VUmc for external validation. The dataset was unlabelled and included tabular clinical measurements and discharge letters. Silver labels for LVEF classes were generated by combining diagnosis codes, echocardiography results, and textual mentions. Gold labels were manually annotated for 300 patients for testing. Multiple Transformer-based (black-box) and Aug-Linear (white-box) models were trained and compared with baselines on structured and unstructured data. To evaluate interpretability, two clinicians annotated 20 discharge letters by highlighting information they considered relevant for LVEF classification. These were compared to SHAP and LIME explanations from black-box models and the inherent explanations of Aug-Linear models.\\
\textbf{Results:} BERT-based and Aug-Linear models, using discharge letters alone, achieved the highest classification results (AUC=0.84 for BERT, 0.81 for Aug-Linear on external validation), outperforming baselines. Aug-Linear explanations aligned more closely with clinicians’ explanations (Cohen's Kappa=$0.25\pm0.07$, Krippendorff's Alpha=$0.21\pm0.09$, Kendall's Tau=$0.23\pm0.07$), than post-hoc explanations on black-box models (Cohen's Kappa=$0.11\pm0.01$, Krippendorff's Alpha=$0.05\pm0.05$, Kendall's Tau=$0.05\pm0.06$).\\
\textbf{Conclusions:} Discharge letters emerged as the most informative source for phenotyping HF patients. Aug-Linear models matched black-box performance while providing clinician-aligned interpretability, supporting their use in transparent clinical decision-making.
}
\vspace{1em}
\\
\\
\noindent\textbf{Keywords:} Natural Language Processing, Discharge letters, Interpretability, Heart Failure

% \boxedtext{
% \begin{itemize}
% \item Key boxed text here.
% \item Key boxed text here.
% \item Key boxed text here.
% \end{itemize}}

%TC:endignore

\section{Introduction}
\label{sec:intro}
Heart Failure (HF) is a chronic disease characterized by the heart's inability to adequately supply blood to the body. It affects 1–2\% of the adult population and over 10\% of the elderly~\cite{mcdonagh20212021}, with a five-year mortality rate of 50\% and frequent hospitalizations~\cite{groenewegen2020epidemiology}. Effective treatment relies on precise phenotyping, but HF’s diverse etiologies and symptoms make this challenging.
Accurate phenotyping of HF patients using Electronic Health Record (EHR) data can enhance clinical decision-making and reduce mortality. However, much of the relevant information is embedded in unstructured text, requiring Natural Language Processing (NLP) techniques for automated extraction~\cite{tayefi2021challenges, cowie2017electronic}.

A key classification parameter for HF patients is the Left Ventricular Ejection Fraction (LVEF)~\cite{mcdonagh20212021}, a numerical value measured using echocardiography, which is used to classify patients into three classes: reduced (HFrEF), mildly reduced (HFmrEF) and preserved (HFpEF) ejection fraction. These classes are an important parameter to guide the treatment of HF patients~\cite{mcdonagh20212021}.
However, while LVEF values may not always be available as structured data, LVEF class can often be recognized from the information reported in clinical texts.

Automatically inferring LVEF class can support clinicians in managing hospitalized HF patients when echocardiographic data are unavailable or delayed, and aid researchers in defining cohorts that require LVEF classes.

In this work, we propose and compare multiple NLP models for phenotyping HF patients in LVEF classes using discharge letters, focusing on distinguishing between HFrEF and HFpEF. Our cohort includes HF patients hospitalized at Amsterdam UMC (locations AMC and VUmc) between 2015 and 2023, covering 16,334 patients with 33,105 hospitalizations. We train models with data from AMC patients, and externally validate on VUmc patients. 
Given the limited amount of manually labelled data,
we propose a strategy to derive silver labels from various sources, including diagnosis codes, echocardiography results and textual mentions.

We compare black-box large language models (LLMs), including encoder-only (i.e., BERT-based~\cite{kenton2019bert}) and decoder-only (e.g., Mistral~\cite{jiang2023mistral}) Transformer models, with \textit{inherently interpretable} linear models augmented with BERT embeddings (e.g., Aug-Linear~\cite{singh2023augmenting}). Performance of models using unstructured data are compared to baselines utilizing structured data.
We evaluate classification performance and assess the interpretability of the different models by contrasting the direct interpretation given by inherently interpretable models with post-hoc explanation methods widely used to interpret black-box models, such as LIME~\cite{ribeiro2016should} and SHAP~\cite{lundberg2017unified}. For this comparison, two clinicians (MCV and DK) manually annotated a subset of data highlighting parts of the text they deemed clinically relevant for the classification.

Our main contributions are:
\begin{itemize}
    \item We introduce a strategy to develop classification models for LVEF classes in absence of explicit mentions of LVEF values and under limited amounts of labelled data.
    \item We propose the first model to phenotype HF patients from Dutch discharge letters, improving classification results with respect to the state-of-the-art models (that uses structured data).
    \item To the best of our knowledge, we conduct the first in-depth analysis of the interpretability provided by Aug-Linear models compared to post-hoc explanations of black-box BERT-based models in the medical domain.
\end{itemize}
\section{Background}
\label{sec:background}
\subsection{HF classification} Numerous studies have investigated the characteristics of HF patients with reduced or preserved ejection fraction, encompassing symptoms, comorbidities, pathophysiology, and treatments~\cite{mentz2014noncardiac, tromp2018identifying, mele2018left, packer2021differential, lauritsen2018characteristics}.
In particular, several models predict HF classes using structured EHR data~\cite{uijl2020registry, desai2021epidemiologic, sepehrvand2023predicting}. 
%In particular, in \cite{uijl2020registry}, the authors proposed different logistic regression-based models to distinguish between reduced and preserved EF using different subsets of covariates. 
%We use the model without NT-proBNP and NYHA and threshold at 40\% as our reference baseline for classification based solely on structured data.

\subsection{Medical NLP} Information extraction from unstructured data utilizing NLP-techniques is a growing trend in the medical domain. Initially dominated by rule-based approaches~\cite{sheikhalishahi2019natural}, the field has seen the emergence of deep learning-based methods~\cite{hao2021health}, in particular with the advent of the first domain-specific transformer-based models, such as BioBERT~\cite{lee2020biobert}. While most work focused on English medical documents, other languages have received increased attention in recent years~\cite{neveol2018clinical}.
In particular, several studies have applied and developed rule-based~\cite{nobel2020natural}, recurrent neural network models~\cite{bagheri2020automatic, sammani2021automatic}, traditional machine learning models~\cite{seinen2023added, dormosh2023predicting}, and, more recently, Transformer-based models~\cite{homburg2023natural} for Dutch clinical documents.
Notably, MedRoBERTa.nl is a RoBERTa-based model for Dutch clinical documents that is publicly available~\cite{verkijk2021medroberta}.

\subsection{Application of NLP for HF classification} Textual data for HF patients have also been analyzed in multiple studies, predominantly focusing on the identification of the diagnosis of HF and its symptoms~\cite{kaspar2018underestimated, moore2021ascertaining, li2022kti} as well as predicting (re)hospitalizations~\cite{liu2019predicting, ambrosy2021natural}.  
While some studies aimed to assess LVEF classes from clinical documents~\cite{garvin2012automated, kim2017extraction, wagholikar2018extraction}, they are limited to extracting explicit mentions of LVEF. In contrast, our work proposes a classification model for LVEF classes from clinical documents \textit{in the absence of mentions of LVEF values}.

\subsection{Interpretability} A core aspect of this study is interpretability, which can be defined as the extent to which we can predict what the model will do, given a change in the input~\cite{miller2019explanation}. Traditional models like logistic regression and rule-based NLP techniques are considered inherently interpretable, while various studies applied post-hoc explanation techniques on black-box models for structured data, such as SHAP or partial dependency plots~\cite{rao2022explainable, ren2023predicting, tasnim2023explainable}. A few studies have applied these explainability techniques to black-box NLP models, focusing on SHAP, LIME and the neural network attention mechanism, both in cardiology~\cite{martin2023hypertension, lovelace2019explainable, lahlou2021explainable} and in other medical domains~\cite{diao2021automated, dolk2022evaluation, blanco2022implementation, li2023interpretable}. 
However, post-hoc explanations have known limitations that are well documented in the literature, such as a lack of faithfulness and proneness to confirmation bias~\cite{jacovi-goldberg-2020-towards, HUANG2024109112, cina2023fixing, wiegreffe2019attention}.

In the current study, we propose the use of the Aug-Linear model presented in~\cite{singh2023augmenting}---which embeds $n$-grams with BERT and uses those in a generalized linear model--- for model interpretation. To the best of our knowledge, this model has never been applied to the clinical domain, nor has the quality of its explanations been evaluated using domain knowledge. 

\section{Materials and Methods}
\label{sec:mat-methods}

\begin{figure}
    \centering
    \includegraphics[height=.9\textheight]{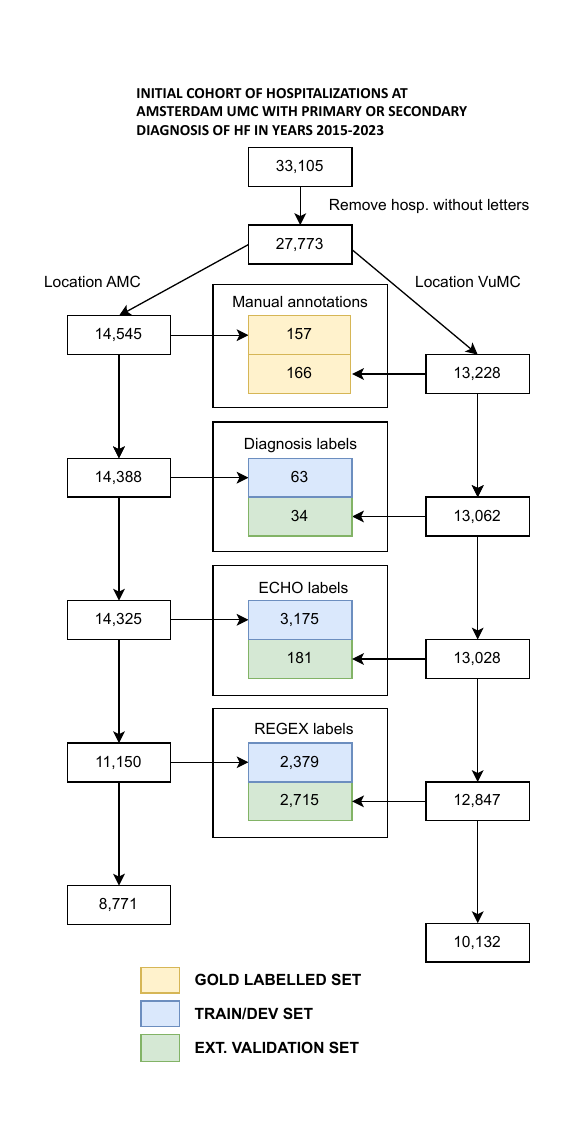}
    \caption{Diagram detailing the different labelling of hospitalizations and the definition of gold, training/dev and external test set}
    \label{fig:data}
\end{figure}

\subsection{Data}
\label{subsec:data}
In this section, we describe the data used in the study and the labelling procedures for classification and interpretability evaluation.
\subsubsection{Dataset}
The data analyzed in this study comprises hospitalizations at Amsterdam UMC, including locations AMC and VUmc, between 2015 and 2023 with a primary or secondary diagnosis of heart failure, totalling 33,105 records of 16,334 unique patients. Each hospitalization record includes demographic information, vital signs, laboratory results, primary and secondary diagnoses, past medical history, echocardiography results, and discharge letters. The study was performed in accordance with the Declaration of Helsinki, and it was approved by the local institutional ethics review board (METC Amsterdam UMC, protocol nr. 2023.0154). 
See Appendix \ref{app:A} for the list of ICD-10-CM codes used for selecting the cohort and for a description of the structured data. 

First, we follow the ESC guidelines~\cite{mcdonagh20212021} and define HFrEF as LVEF $<40\%$ and HFpEF as LVEF $>50\%$.
A large part of our dataset lacks gold-standard labels to distinguish between HFrEF and HFpEF patients. This is because, while ICD-10-CM codes exist to indicate these specific characteristics, physicians primarily use generic HF ICD-10-CM codes. As a result, we derive silver labels for HFrEF and HFpEF by utilizing various sources of information.
We divide the dataset into subsets for model training, testing, and external validation, based on the source of the derived labels (see Figure~\ref{fig:data}). In particular, we reserve hospitalizations from VUmc hospital for external validation and use those from AMC hospital for model training. 

\subsubsection{Gold and silver labelling for classification}
After excluding hospitalizations for which a discharge letter was not available, 300 patients were randomly selected to have their hospitalizations manually annotated by MB to form our gold-standard test set for classification evaluation.
For the remaining data, used for model training and external validation, we derive silver labels. The first sources of silver labels are medical diagnosis code tables, including a combination of ICD-10-CM codes and SNOMED-CT codes, some of which specify HFrEF or HFpEF. 
The second source is echocardiographic results, which include measurements of the LVEF.
Since LVEF might show improvement due to treatment, we link each hospitalization to all echocardiographic measurements from the same patient within a 90-day window before admission and after discharge. If any of the reports during this period includes a measured LVEF $<40\%$, we assign the HFrEF label to the case. If no report with LVEF $<40\%$ exists, but there is at least one with LVEF $\geq 50\%$, we assign an HFpEF label. Otherwise, no echocardiography-based label is assigned.
Although structured, these data are less certain than diagnostic codes, as echocardiographic measurements can vary depending on the method used and medication effects. 
When neither codes nor echocardiography results provide silver labels, we analyze the text of discharge letters. In some cases, they contain explicit mentions of LVEF values, and we extract such information using regular expressions to derive silver labels. See Appendix \ref{app:B} for more details on gold and silver labelling. To assess if our silver labels reflect inaccuracies or biases inherent to the data, we examine whether their missingness is completely at random (MCAR), at random (MAR) or not at random (MNAR) by predicting it with a logistic regression on structured variables.

\begin{figure}
    \centering
    \includegraphics[width=.8\textwidth]{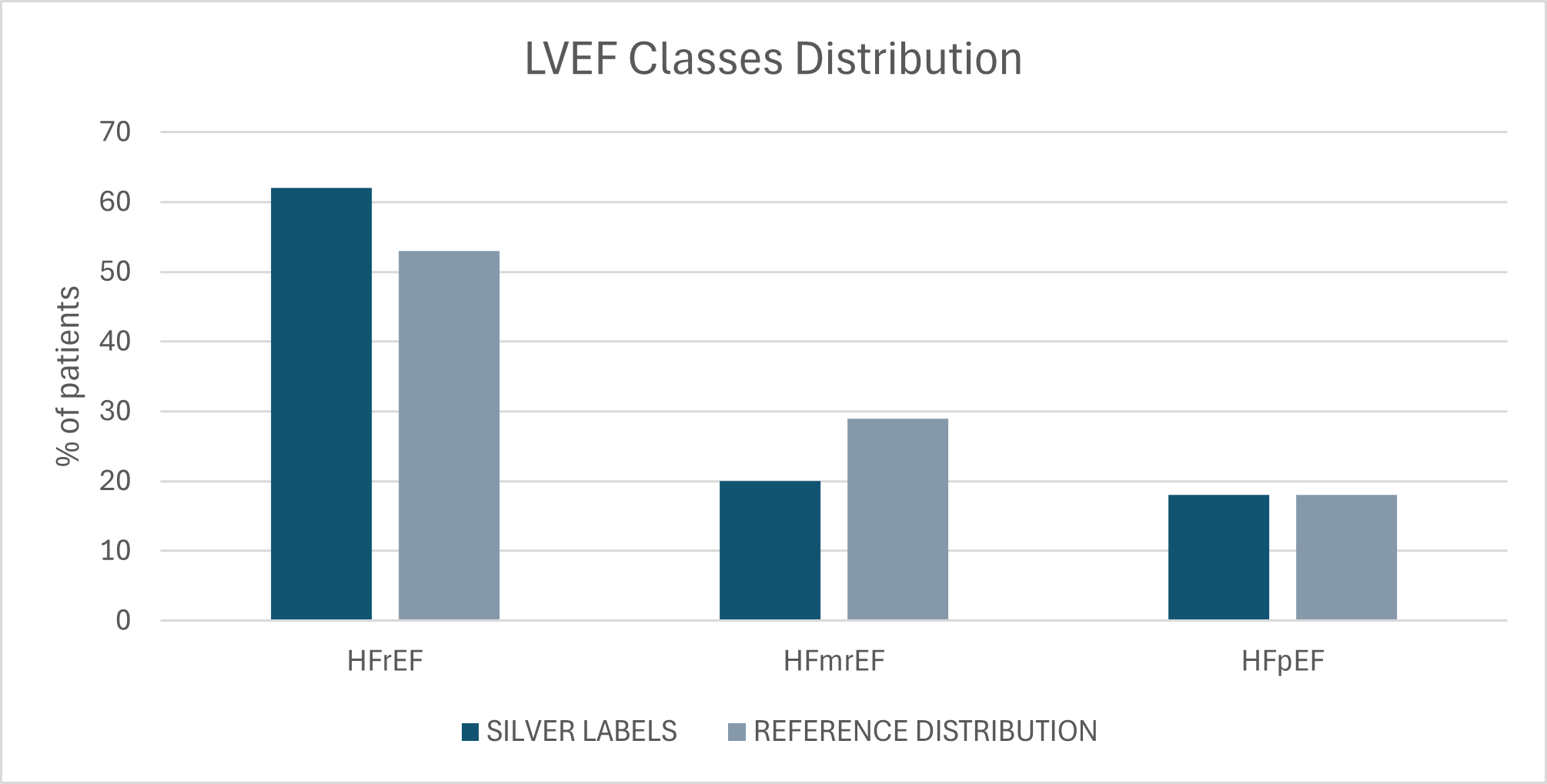}
    \caption{LVEF classes in our data (silver labels) vs. reference distribution from \cite{kaplon2022comprehensive}.}
    \label{fig:data-labels}
\end{figure}

\begin{figure}
    \centering
    \includegraphics[width=\linewidth]{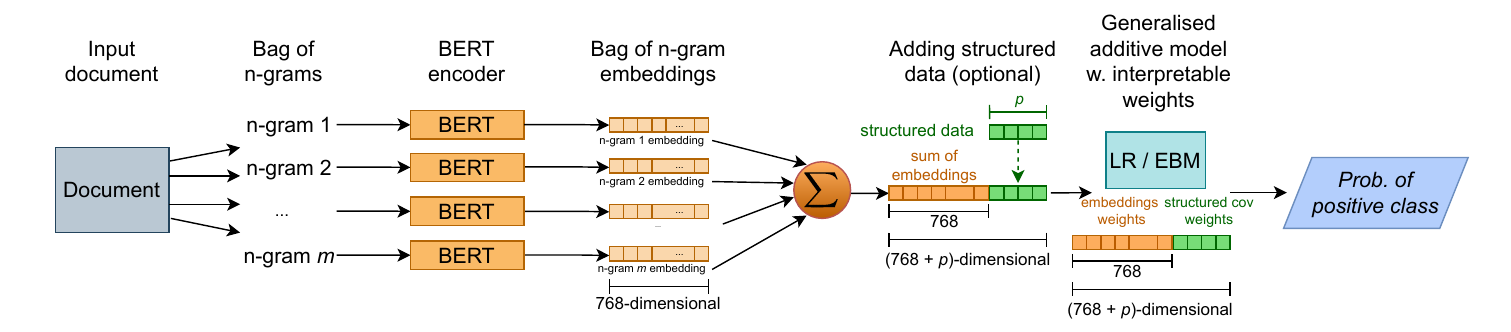}
    \caption{Schema of the training procedure for Aug-Linear models, including the optional addition of structured covariates.}
    \label{fig:auggam-schema}
\end{figure}

\subsubsection{Gold labelling for interpretability}
To assess and compare the interpretability of our models, two clinicians (MCV and DK) manually annotated 20 discharge letters, highlighting what they considered relevant to classify the patients.
First, $n$-grams in each discharge letter are annotated in how much they disclose the patient's LVEF class 
%as HFrEF vs HFpEF  
(local explanations).
Given a letter and its correct label (HFrEF or HFpEF), the clinician annotates how certainly the $n$-gram is related to the label (without doubt, strong indication, $n$-gram is an indication for the opposite label). The annotation process followed specific guidelines developed iteratively to ensure consistency and accuracy. A detailed description of the annotation procedure is provided in Appendix \ref{app:E}.
Subsequently, the same clinicians also assign a binary label (relevant/not relevant) to the top-15 most important $n$-grams produced by the different models (global explanations). 

\subsection{Classification Models}
\label{subsec:models}
We use classification models based on \textit{only structured data}, \textit{only unstructured data}, or \textit{both structured and unstructured data}.
Please refer to Appendix \ref{app:C} for more details.

\subsubsection{Classification from structured data}
Our main baseline using structured covariates is the logistic regression (LR) model presented in~\cite{uijl2020registry}. 
We use the version that excludes NYHA class and NT-proBNP---due to the unavailability of these variables for the majority of our patients—--which results in 20 structured input variables. We compare the two different LVEF thresholds they used (40 for \uijletalorig{40} and 50 for \uijletalorig{50}).
We also retrain the LR model from~\cite{uijl2020registry} on our data using the same 20 variables (\uijletalstruct).
Finally, we train an Explainable Boosting Machine (EBM) model~\cite{lou2013accurate} using the same covariates (\ebmstruct).

\subsubsection{Classification from discharge letters}
When using only unstructured texts without structured covariates, we use a LR and an EBM classifier on a TF-IDF representation~\cite{leskovec2020mining} of the discharge letters as baselines (\lrtfidf{} and \ebmtfidf{}, respectively).
We also use three black-box models for document classification: \robbert{}~\cite{delobelle2020robbert}, \medrobertanl{}~\cite{verkijk2021medroberta}, and \geitje{}~\cite{vanroy2023language}. Due to the 512-token input limit of RobBERT and MedRoBERTa.nl, we split the letters into 512-token chunks, using the maximum probability across splits for classification.

We compare these three models with Aug-Linear~\cite{singh2023augmenting}, whose training process is illustrated in Figure \ref{fig:auggam-schema}.
Fitting Aug-Linear models has two steps: 1) extract embeddings for an input, and 2) use these embeddings to fit an interpretable model (i.e., a linear model).
For step 1, we extract $n$-grams for our discharge letters and embed each $n$-gram independently with our best Transformer-based black-box model.
For step 2, we use these $n$-gram embeddings to train two inherently interpretable models: a LR (\auggamlr) and an EBM classifier (\auggamebm).

\subsubsection{Classification from structured data and discharge letters}
Finally, we also train our Aug-Linear models on the combination of structured and unstructured variables by directly concatenating the structured covariates used in our baselines to the $n$-gram feature representations learned by Aug-Linear with LR and EBM (Figure \ref{fig:auggam-schema}).
We refer to these models as \auggamlrstruct{} and \auggamebmstruct{}.

\subsection{Explainability methods}
\label{subsec:explainability}
To assess the interpretability of our models for textual data, we compare post-hoc explanation techniques for black-box models with the interpretation provided by the Aug-Linear models.
Post-hoc attribution methods are widely used in the literature and are very relevant since most state-of-the-art models currently in use are black-boxes.

In this work, we choose LIME~\cite{ribeiro2016should} and SHAP~\cite{lundberg2017unified} as post-hoc explanation techniques since they are among the most commonly used techniques with NLP classifiers. 
Details about explainability methods' implementations are available in Appendix \ref{app:D}.

\subsubsection{Local and global explanations}
Local explanations are explanations of a model for a specific input.
Aug-Linear models assign a score for each $n$-gram in the input, computed by multiplying the $n$-gram embedding vector with the parameter vector of the linear model (see Appendix~\ref{app:auggam}).
LIME and SHAP compute token-level scores, i.e., token contributions to the positive or negative class. 

Global explanations are explanations of a model \textit{in general}, i.e., for any possible input.
Aug-Linear models have a proper global explanation since each $n$-gram can compute its contributions independently of a specific input.
This makes these models attractive since they can be inspected or even ``debugged'' \textit{globally}~\cite{singh2023augmenting}.
Though LIME and SHAP do not compute proper global explanations, we follow~\cite{christoph2020interpretable} and approximate them by computing explanations on a test set and averaging the per-token scores.

\subsubsection{Reliance on $n$-gram frequency for predicting outcomes}

For each method, we compute the top-50 relevant $n$-grams for each class.
To verify how much different models rely on $n$-gram frequency to correlate covariates to outcomes, we compute the following score $s$ for each model:

$$s = \sum_{i \in \mathcal{T}_\text{true}} (e_i \cdot c_i) - \sum_{i \in \mathcal{T}_\text{false}} (e_i \cdot c_i),$$
\noindent
where $\mathcal{T}_\text{true}$ ($\mathcal{T}_\text{false}$) is the set of the top 50 most relevant $n$-grams for the positive (negative) class,
$e_i$ is the explanation score
for $n$-gram $i$ and $c_i$ is the frequency of $n$-gram $i$ in the samples of the class.
The higher this score, the more the model relies on $n$-grams frequencies.

\subsection{Training procedure and evaluation}
\label{subsec:training}
\subsubsection{Training procedure}
All models except \uijletalorig{40}, \uijletalorig{50}, and \geitje{} are trained using 10-fold stratified cross-validation (CV).
Hyperparameters are selected via grid search.

Since our silver labels can be partially derived from the content in the letters, for models using text data we masked LVEF expressions in the training set, while we kept these expressions in the test set to allow for an evaluation in a realistic setting where this type of information can be present.

In \geitje{} we use the text of the letter as input preceded by the prompt \textit{``U bent cardioloog en bekijkt een ontslagbrief van een patiënt met hartfalen. Antwoord “Systolisch” of “Diastolisch”, afhankelijk van het type hartfalen. Tekst:"} (\textit{``You are a cardiologist reviewing a discharge letter from a patient with heart failure. Answer “Systolic” or “Diastolic,” depending on the type of heart failure. Text:"}). We then parse the output, checking if it corresponds to \textit{“Systolisch"} (HFrEF) or \textit{“Diastolisch"} (HFpEF). If this is not the case, we repeat the execution.
If the model does not produce a valid output after 10 iterations, we judge the sample as incorrectly classified.

For \auggamlr{} and \auggamebm{}, we train models including progressively higher-order $n$-grams---where $n \in [1,5]$--- starting with models trained on unigrams, then unigrams and bigrams, and so on.
Before extracting $n$-grams, we replace numbers with a placeholder and remove punctuation.
The $n$-grams with lower frequencies are removed, with a threshold for each $n$ selected via grid search.

\subsubsection{Classification evaluation}
As metrics for the classification task, we compute the area under the receiver operating characteristic curve (AUC), precision (P), recall (R), and F1-score. P, R, and F1 are computed using the classification optimal threshold as defined by Youden's index~\cite{youden1950index}.

Results on AMC hospital silver-labelled data are used for model and hyperparameter selection, while the models are eventually evaluated on the gold-labelled test set and on the silver-labelled external validation set from VUmc hospital.

\subsubsection{Explanation evaluation}
We evaluate explainability methods at both global and local levels.
For local explanations, we compute the agreement between the ground truth explanations derived by manual annotations and the explanations produced by Aug-Linear models, LIME and SHAP. The agreement is computed via Cohen's Kappa~\cite{mchugh2012interrater}, Krippendorff's Alpha~\cite{krippendorff2011computing}, F1-score and Kendall's Tau~\cite{kendall1938new}. 
For global explanations, we evaluate the number of \textit{relevant} n-grams marked by annotators in the global explanations from each method. More details are in Appendix~\ref{app:expl-eval}.

Since LIME and SHAP have explanations at the token level, we compare them to both unigram-based Aug-Linear models and our best Aug-Linear models.
\section{Results}
\label{sec:results}

\begin{table}[]
\centering
\small
\caption{Classification results on gold-labelled dataset and on the external validation dataset. P = precision. R = recall. F1 = F1 score. AUC = Area under the receiver operating characteristic curve. We show results for models that use structured data only, discharge notes only (baselines using TF-IDF representations, black-box, and white-box models, respectively), and models that combine structured and unstructured data.}
\label{tab:res-ext}
\resizebox{\textwidth}{!}{\begin{tabular}{@{}lr@{\hskip 0.2in}rrrr@{\hskip 0.2in}rrrr@{}}
\toprule
&& \multicolumn{4}{c}{Gold-labelled dataset} & \multicolumn{4}{c}{External validation dataset} \\ %\midrule
&  Model & P {[}\%{]} & R {[}\%{]} & F1 {[}\%{]} & AUC {[}\%{]} & P {[}\%{]} & R {[}\%{]} & F1 {[}\%{]} & AUC {[}\%{]} \\ \midrule
\multirow{3}{*}{\rotatebox[origin=c]{90}{\parbox[c]{1cm}{\centering Struct. data}}}
& \uijletalorig{40}    & 81.73 & 57.43  & 67.46 & 56.67 & 63.41 & 62.59 & 63.00 & 67.89 \\
& \uijletalorig{50}    & 80.00 & 45.95  & 58.37 & 53.64 & 64.52 & 65.88 & 65.19 & 68.91 \\
& \uijletalstruct{}    & 82.98 & 52.70  & 64.46 & 54.52 & 66.44 & 65.78 & 66.11 & 74.05 \\
& \ebmstruct{}         & 84.21 & 54.05  & 65.84 & 55.01 & 73.56 & 69.12 & 71.27 & 75.66 \\ \midrule
\multirow{3}{*}{\rotatebox[origin=c]{90}{\parbox[c]{3cm}{\centering Unstructured data (discharge letters)}}}
& \lrtfidf{}           & 82.52 & 57.43 & 67.73 & 58.65 & 61.78 & 71.98 & 66.49 & 74.02 \\
& \ebmtfidf{}          & 82.83 & 55.41 & 66.40 & 57.41 & 63.52 & 68.44 & 65.89 & 71.32 \\
\cmidrule{3-10}
& \medrobertanl{}      & \textbf{92.73} & \textbf{68.92} & \textbf{79.07} & \textbf{73.17} & \textbf{84.44} & 74.98 & \textbf{80.15} & \textbf{83.52} \\
& \robbert{}           & 89.22 & 61.49 & 72.80 & 65.44 & 77.45 & \textbf{82.31} & 79.81 & 78.55 \\
& \geitje{}            & 89.22 & 61.49 & 72.80 & -     & 76.51 & 72.41 & 74.40 & - \\
\cmidrule{3-10}
& \auggamlr{}          & 91.35 & 64.19 & 75.40 & 68.54 & 74.01 & 73.36 & 73.68 & 80.77 \\
& \auggamebm{}         & 91.18 & 62.84 & 74.40 & 67.56 & 72.57 & 79.97 & 75.12 & 80.10 \\ \midrule
\multirow{2}{*}{\rotatebox[origin=c]{90}{\parbox[c]{0.9cm}{\centering Both}}}
& \auggamlrstruct{}    & 90.00 & 60.81 & 72.58 & 66.11 & 73.12 & 72.45 & 72.78 & 80.12 \\
& \auggamlrstruct{}    & 89.69 & 58.78 & 71.02 & 62.12 & 71.10 & 72.54 & 71.81 & 80.35 \\
\bottomrule
\end{tabular}}
\end{table}

A total of 27,773 cases were included in the full analysis, 97 have been assigned a silver label using medical diagnosis codes, 3,356 using echocardiography and 5,094 via free text search (Figure~\ref{fig:data}). The silver labels are not MCAR since the LR model for missingness on structured variables has AUC of 0.68. In Figure~\ref{fig:data-labels}, the distribution of our silver labels and a reference distribution including $5,000$ hospitalized HF patients from 33 countries~\cite{kaplon2022comprehensive} are compared. The Jensen-Shannon divergence is very small ($0.0085$), suggesting that our silver labels are MAR.

Table \ref{tab:res-ext} reports classification results in terms of precision, recall, F1 score and AUC on the manually annotated set and on the external validation set for the different models we developed and tested. Additional results, considering different hyperparameters, are reported in Appendix \ref{app:C}. 

Models based only on structured data and interpretable TF-IDF models reach similar performance, and black-box language models outperform both, with MedRoBERTa.nl achieving the best result (AUC=0.84 on external validation). The GEITje LLM overcome simpler models but not the BERT-based ones. The Aug-Linear models are able to achieve an AUC of 0.81 on the external validation set, which is near the best black-box models. Adding structured data to them does not improve the performance. Results on gold-labelled set are lower, but aligned in the models ranking and distances.

Table \ref{tab:freq-score} shows the computation of the frequency score for each model. The TF-IDF-based models are those with the highest scores, meaning these models are the ones that rely most heavily on $n$-gram frequencies in their predictions. Notably, the Aug-Linear models are positioned between BERT-based (less reliant on frequencies) and TF-IDF models.

\begin{table}[]
\centering
\caption{$n$-gram frequency score per model. The lower the frequency score, the less the model relies on $n$-gram frequencies.}
\label{tab:freq-score}
\begin{tabular}{@{}lllr@{}}
\toprule
\makecell[l]{Model \\ Backbone} & \makecell[l]{Model Training} & Interpretability & \makecell{Frequency Score ($\downarrow$)}\\
\midrule
MedRoberta.nl   & \multirow{2}{*}{End-to-end}    & SHAP & 3.24 \\
RobBERT         &     & SHAP & 4.25 \\
\cmidrule{2-4}
MedRoberta.nl   & \multirow{3}{*}{LR}            & Intrinsic & 10.15 \\
RobBERT         &             & Intrinsic & 9.84 \\
TF-IDF          &             & Intrinsic & 17.55 \\
\cmidrule{2-4}
MedRoberta.nl   & \multirow{3}{*}{EBM}           & Intrinsic & 9.54 \\
RobBERT         &            & Intrinsic & 9.66 \\
TF-IDF          &            & Intrinsic & 18.99 \\
\bottomrule
\end{tabular}
\end{table}

\begin{figure}
    \centering
    \includegraphics[height=.9\textheight]{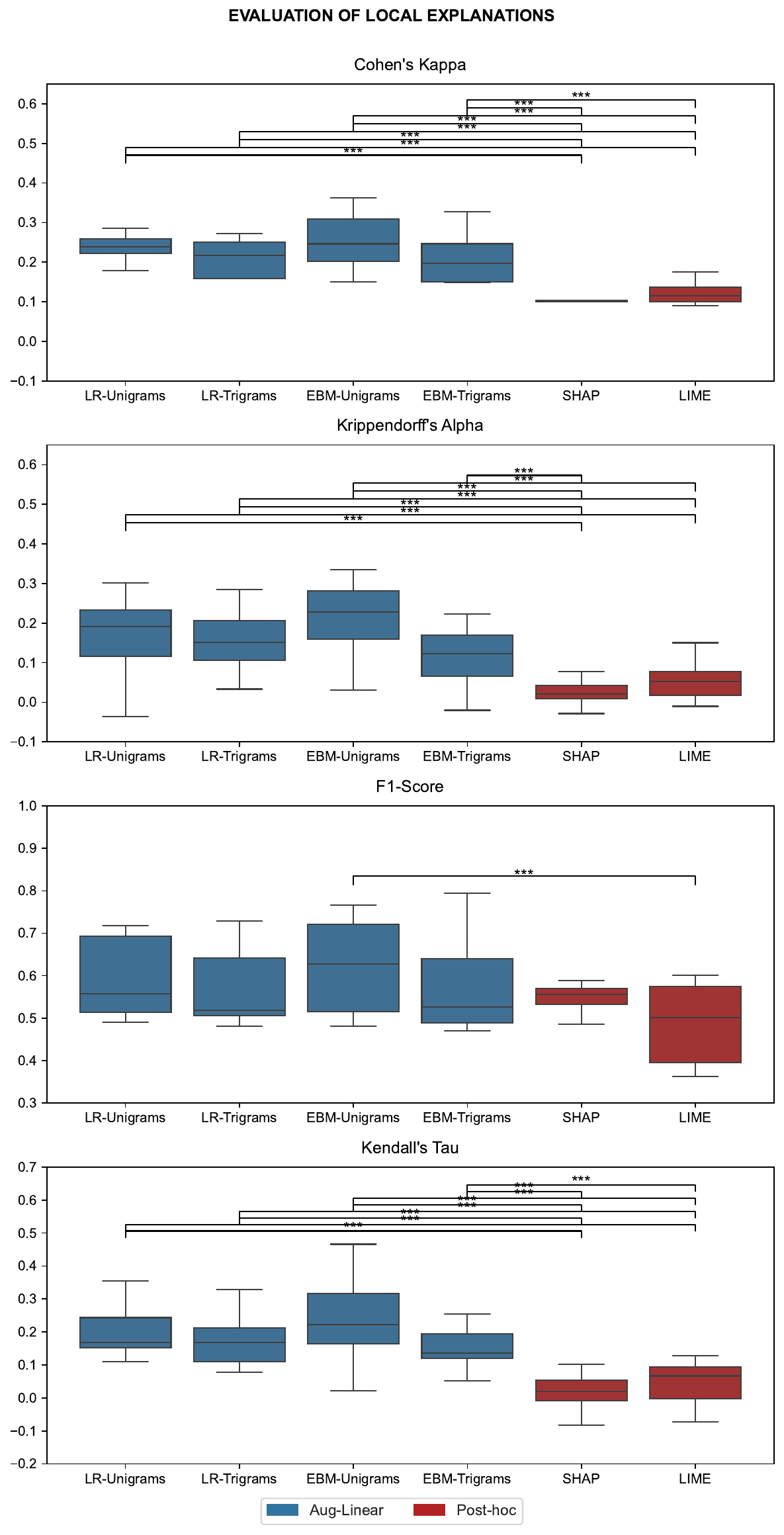}
    \caption{Results for the evaluation of local explanations, computing agreement between the different explanation methods and the manual annotations, considering three tags: no indication, indication for the correct class, indication for the incorrect class. P-values of Mann-Whitney U test for differences in medians with Bonferroni correction: $*** < 0.001$}
    \label{fig:local-expl-eval}
\end{figure}

\begin{figure}
    \centering
    \includegraphics[width=0.9\textwidth]{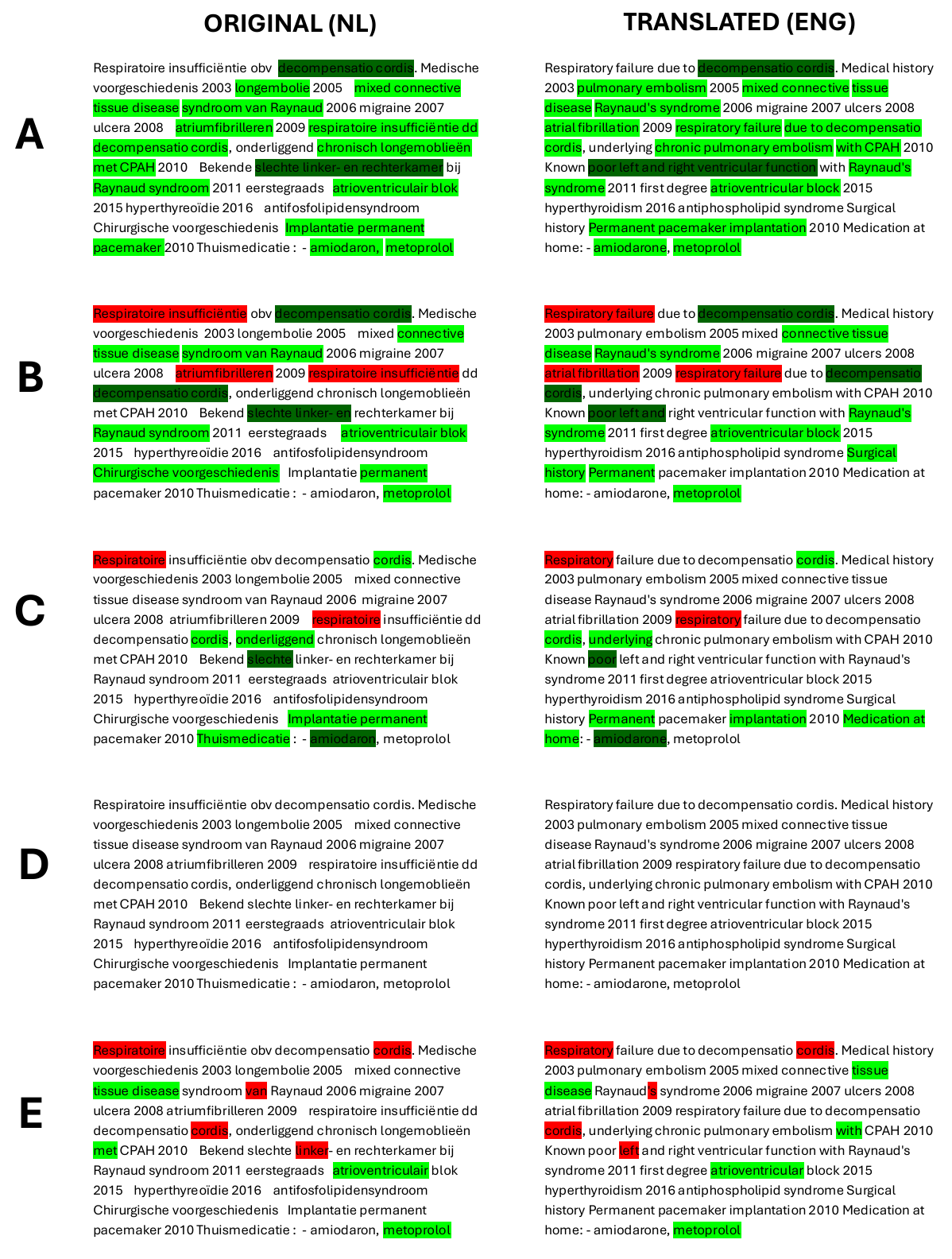}
    \caption{Example of local explanations on a chunk of a (fictitious) discharge letter for a HFrEF patient, with different methods. A. Manual Annotations from clinicians, B. EBM Aug-Linear with trigrams, C. LR Aug-Linear with unigrams D. LIME on MedRoberta.nl, E. SHAP on MedRoberta.nl. Dark Green = Complete giveaway indication for HFrEF, Green = Strong indication for HFrEF, Red = Indication for HFpEF.}
    \label{fig:expl-example}
\end{figure}

Figure~\ref{fig:local-expl-eval} summarizes the alignment of the local explanations with the manual annotations. For the majority of the metrics, there are significant differences between the alignment obtained by the Aug-Linear models and the ones achieved by SHAP and LIME. An example is reported in Figure \ref{fig:expl-example}. 

Table \ref{tab:global-summary} summarizes the evaluation of the global explanations, with Table~\ref{tab:global-expl-auglinear-tri} reporting the most relevant n-grams highlighted by the Aug-Linear models with trigrams, along with their evaluation by annotators. 

Additional results on interpretability are reported in Appendix~\ref{app:D}.

\begin{table*}[]
\centering
\caption{Manual evaluation of global explanations, as number and percentage of n-grams marked as relevant for each class and on average}
\label{tab:global-summary}
\begin{tabular}{@{}lrrrrrr@{}}
\toprule
Model & \multicolumn{1}{l}{HFrEF \#} & HFrEF \% & \multicolumn{1}{l}{HFpEF \#} & HFpEF \% & \multicolumn{1}{l}{Average \#} & \multicolumn{1}{l}{Average \%} \\ \midrule
LR-Trigrams  & 7  & 46.67  & \textbf{6} & \textbf{40.00} & 6.5 & 43.33 \\
EBM-Trigrams & \textbf{15} & \textbf{100.00} & 2 & 13.33 & \textbf{8.5} & \textbf{56.57} \\
LIME         & 2  & 13.33  & 3 & 20.00 & 2.5 & 16.67 \\
SHAP         & 1  & 6.67   & 0 & 0.00  & 0.5 & 3.33  \\
LR-Unigrams  & 4  & 26.67  & 4 & 26.67 & 4.0 & 26.67 \\
EBM-Unigrams & 3  & 20.00  & 2 & 13.33 & 2.5 & 16.67 \\ \bottomrule
\end{tabular}
\end{table*}

\begin{table*}[]
\centering
\caption{Global explanations as the 15 most relevant n-grams for HFrEF (upper part) and HFpEF (lower part) for Aug-Linear models with trigrams. Green backgrounds are those assessed as clinically relevant.}
\label{tab:global-expl-auglinear-tri}
\scriptsize
\begin{tabular}{lp{3.5cm}p{3.5cm}p{3.5cm}p{3.5cm}}
\toprule
 \multicolumn{1}{c}{\multirow{2}{*}{\#}} &
  \multicolumn{2}{c}{\textbf{AUG-Linear LR TRI (HFrEF)}} &
  \multicolumn{2}{c}{\textbf{AUG-Linear EBM TRI (HFrEF)}} \\
\multicolumn{1}{c}{} &
  \multicolumn{1}{c}{N-GRAM {[}NL{]}} &
  \multicolumn{1}{c}{N-GRAM {[}ENG{]}} &
  \multicolumn{1}{c}{N-GRAM {[}NL{]}} &
  \multicolumn{1}{c}{N-GRAM {[}ENG{]}} \\ \midrule
1 &
  \cellcolor[HTML]{8ED973}mellitus hypercholesterolemie &
  \cellcolor[HTML]{8ED973}mellitus hypercholesterolemia &
  {\cellcolor[HTML]{8ED973}slechte linkerventrikelfunctie ejectiefractie} &
  \cellcolor[HTML]{8ED973}poor left ventricular function ejection fraction \\
2 &
  levenslang ticagrelor &
  ticagrelor for life &
  {\cellcolor[HTML]{8ED973}slechte tot matige} &
  \cellcolor[HTML]{8ED973}poor to moderate \\
3 &
  onderzoek oesofagogastro-duodenoscopie &
  examination oesofagogastro-duodenoscopy &
  {\cellcolor[HTML]{8ED973}slecht tot matige} &
  \cellcolor[HTML]{8ED973}poor to moderate \\
4 &
 een naaste op &
  a neighbour at &
  {\cellcolor[HTML]{8ED973}matig tot slechte} &
  \cellcolor[HTML]{8ED973}moderate to poor \\
5 &
  laatst innemen &
  last take &
  {\cellcolor[HTML]{8ED973}cardiomyopathie met matigslechte} &
  \cellcolor[HTML]{8ED973}cardiomyopathy with moderate-severe \\
6 &
  \cellcolor[HTML]{8ED973}matig ernstige &
  \cellcolor[HTML]{8ED973}moderately severe &
  {\cellcolor[HTML]{8ED973}matig tot slecht} &
  \cellcolor[HTML]{8ED973}moderate to poor \\
7 &
  \cellcolor[HTML]{8ED973}matig tot slechte &
  \cellcolor[HTML]{8ED973}moderate to poor &
  {\cellcolor[HTML]{8ED973}cardiomyopathie met matigredelijke} &
  \cellcolor[HTML]{8ED973}cardiomyopathy with moderate \\
8 &
  cardiologie opnamedag &
  cardiology admission day &
  {\cellcolor[HTML]{8ED973}gering tot matige} &
  \cellcolor[HTML]{8ED973}poor to moderate \\
9 &
  \cellcolor[HTML]{8ED973}cardiomyopathie met matigslechte &
  \cellcolor[HTML]{8ED973}cardiomyopathy with moderate-severe &
  {\cellcolor[HTML]{8ED973}matige tot slechte} &
  \cellcolor[HTML]{8ED973}moderate to poor \\
10 &
  \cellcolor[HTML]{8ED973}matig tot slecht &
  \cellcolor[HTML]{8ED973}moderate to poor &
  {\cellcolor[HTML]{8ED973}gedilateerde slechte linker} &
  \cellcolor[HTML]{8ED973}dilated poor left \\
11 &
  een dotterbehandeling van &
  a dotter treatment of &
  {\cellcolor[HTML]{8ED973}diffuus slechte systolische} &
  \cellcolor[HTML]{8ED973}diffuse poor systolic \\
12 &
  \cellcolor[HTML]{8ED973}hypertensie hypercholesterolaemie &
  \cellcolor[HTML]{8ED973}hypertension hypercholesterolaemia &
  {\cellcolor[HTML]{8ED973}geringe tot matige} &
  \cellcolor[HTML]{8ED973}minor to moderate \\
13 &
  \cellcolor[HTML]{8ED973}levenslang carbasalaatcalcium &
  \cellcolor[HTML]{8ED973}lifelong carbasalate calcium &
  {\cellcolor[HTML]{8ED973}cardiomyopathie met slechte} &
  \cellcolor[HTML]{8ED973}cardiomyopathy with poor \\
14 &
  neu opnamedag &
  neu admission day &
  {\cellcolor[HTML]{8ED973}slechte linker} &
  \cellcolor[HTML]{8ED973}poor left \\
15 &
  laatst innemen op &
  last take on &
  {\cellcolor[HTML]{8ED973}matige linkerventrikelfunctie matige} &
  \cellcolor[HTML]{8ED973}{moderate left ventricular function moderate} \\ \midrule
\multicolumn{1}{c}{\multirow{2}{*}{\#}} &
  \multicolumn{2}{c}{\textbf{AUG-Linear LR TRI (HFpEF)}} &
  \multicolumn{2}{c}{\textbf{AUG-Linear EBM TRI (HFpEF)}} \\
\multicolumn{1}{c}{} &
  \multicolumn{1}{c}{N-GRAM {[}NL{]}} &
  \multicolumn{1}{c}{N-GRAM {[}ENG{]}} &
  \multicolumn{1}{c}{N-GRAM {[}NL{]}} &
  \multicolumn{1}{c}{N-GRAM {[}ENG{]}} \\ \midrule
1 &
  \cellcolor[HTML]{8ED973}normale repolarisatie &
  \cellcolor[HTML]{8ED973}normal repolarization &
  van een functionele &
 of a functional \\
2 &
 rejectie behandeld met &
  rejection treated with &
 de hoogte stellen &
  inform \\
3 &
  \cellcolor[HTML]{8ED973}beiderzijds normale &
  \cellcolor[HTML]{8ED973}bilateral normal &
  het weekend en &
  the weekend and \\
4 &
  gevoel bij het &
  feeling at the &
  alat ul &
  alat ul \\
5 &
  \cellcolor[HTML]{8ED973}goede conditie de &
  \cellcolor[HTML]{8ED973}good condition the &
 n meerdere &
 n multiple \\
6 &
  normale densities &
  normal densities &
  {mdo bespreking} &
  multi-disciplinary consultation \\
7 &
  \cellcolor[HTML]{8ED973}respiratoir stabiel &
  \cellcolor[HTML]{8ED973}respiratory stable &
  {tot maart} &
  until March \\
8 &
  normaal sinus &
 normal sinus &
  {\cellcolor[HTML]{8ED973}hypertensie met verhoodge} &
  \cellcolor[HTML]{8ED973}hypertension with elevated \\
9 &
  conclusie ongewijzigde &
  conclusion unchanged &
  {draaien van het} &
  turning it \\
10 &
  \cellcolor[HTML]{8ED973}ejectie fractie &
  \cellcolor[HTML]{8ED973}ejection fraction &
  {s nachts soms} &
  at night sometimes \\
11 &
  eosinofielen &
  eosinophils &
  {dag post implant} &
  day post implant \\
12 &
  \cellcolor[HTML]{8ED973}normaal aspect van &
  \cellcolor[HTML]{8ED973}normal aspect of &
  {pap NUMBER mmhg} &
  pap NUMBER mmhg \\
13 &
  conclusie stabiele &
  conclusion stable &
  {drukpijn of weerstanden} &
  pressure pain or resistances \\
14 &
  \cellcolor[HTML]{8ED973}respiratoir stabiel met &
  \cellcolor[HTML]{8ED973}respiratory stable with &
  {\cellcolor[HTML]{8ED973}dapagliflozine} &
  \cellcolor[HTML]{8ED973}dapagliflozin \\
15 &
  \cellcolor[HTML]{8ED973}conclusie normaal aspect &
  \cellcolor[HTML]{8ED973}conclusion normal aspect &
  {acute biliaire} &
  acute biliary \\ \bottomrule
\end{tabular}
\end{table*}
\section{Discussion}
\label{sec:discussion}

\subsection{Significance of the Problem and Approach}
This study presents the first attempt to classify HF patients using Dutch discharge letters to distinguish between HF with reduced and preserved ejection fraction (HFrEF and HFpEF). Our goal was not only to develop a model for accurate classification, but also to ensure that such classification could be interpreted and validated by clinicians --- a critical requirement for deployment in real-world medical settings. This addresses a dual challenge: the scarcity of structured data such as ICD labels or echocardiography results, and the need for explainable models that support clinical decision-making.

By leveraging free-text discharge letters, we demonstrate that it is possible to recover LVEF classes using NLP methods, with better performance than state-of-the-art models based on structured data alone. While prior work has primarily focused on predictive performance, we show that interpretability does not need to be sacrificed to achieve strong results.

\subsection{Performance and Interpretability Trade-off}
The central aim of this study is to explore how interpretable models can be applied to complex clinical NLP tasks without compromising accuracy. Our results show that Aug-Linear models approach the performance of transformer-based models, reaching an AUC of 80.8\% on the external validation set. Notably, the best-performing Aug-Linear models relied on trigrams, suggesting that most clinically relevant information is captured within short, local spans of text. This reinforces the idea that interpretable models, when properly designed, can extract meaningful patterns from clinical narratives. Results on the gold labelled set are lower but maintain the same model ranking. Its limited sample size and the difficulties in assessing the correct label in some cases might explain this difference.

Although the MedRoBERTa.nl model achieved a slightly higher AUC of 83.5\%, the marginal gain in predictive power comes at the cost of explainability. Our evaluation of interpretability --- based on manual annotations from two clinicians --- showed that post-hoc explanation techniques applied to black-box models often fail to align with clinicians’ reasoning, both at local and global levels. Aug-Linear models consistently produced explanations more aligned with expert annotations, even when restricted to unigrams. At the global level, trigram-based Aug-Linear models outperformed all other methods in identifying class-relevant patterns, particularly for HFrEF.

These findings address a central trade-off in clinical NLP: while black-box models may offer slightly higher accuracy, interpretable models like Aug-Linear are better suited to clinical environments, where transparency and clinician trust are essential for adoption.

Moreover, adding structured data did not improve the performance of any model, suggesting that discharge letters alone --- when analyzed effectively --- contain sufficient information for this classification task.

\subsection{Challenges and Limitations}
Despite the promising results, this study presents some limitations. The use of silver labels --- derived from a combination of structured codes, echocardiography results, and text mentions --- introduces potential label noise that could affect model training and evaluation. Although we evaluated performance on a gold-labelled dataset, it was considerably smaller than the silver-labelled external dataset. Missing silver labels were found to be missing at random, though not completely at random. Additionally, both hospitals involved in this study belong to the same healthcare organization (Amsterdam UMC), which may limit generalizability to other institutions or healthcare systems. Regarding interpretability evaluation, the number of manually reviewed explanations was limited, and the subjective nature of the task led to only fair inter-annotator agreement.

\subsection{Future Work}
Future research will aim to scale and generalize the approach by incorporating data from additional hospitals and regions, ideally with more diverse patient populations and documentation styles. Expanding the manually annotated dataset will also enable a more robust evaluation of both classification performance and interpretability. Additionally, we plan to explore the integration of other types of unstructured data --- such as outpatient visit notes and echocardiography reports --- to assess their contribution to both model accuracy and explainability.

\section{Conclusions}
\label{sec:conclusions}
This study demonstrates that unstructured clinical texts---specifically Dutch discharge letters---can be effectively leveraged to phenotype HF patients by LVEF classes. We show that models based on free-text outperform those using structured data alone, confirming the value of narrative documentation in capturing nuanced clinical information.

More importantly, we highlight the critical role of interpretability in clinical NLP. Our work presents the first comparison between Aug-Linear model explanations and traditional post-hoc methods (SHAP and LIME) in a clinical context, showing that Aug-Linear explanations align more closely with clinicians' reasoning at both local and global levels. These findings support the use of interpretable architectures as viable alternatives to black-box models, particularly in domains like healthcare where trust and transparency are essential.

%%%%%%%%%%%%%%

\section*{Competing interests}
No competing interest is declared.
%\todovittorio{CHECK THIS}

\section*{Author contributions statement}
% CREDIT OPTIONS:
% Conceptualization
% Data curation
% Formal analysis
% Funding acquisition
% Investigation
% Methodology
% Project administration
% Resources
% Software
% Supervision
% Validation
% Visualization
% Writing - original draft
% Writing - review & editing
VT: investigation, software, methodology, writing original draft
MJB: conceptualization, data curation, methodology, writing review \& editing, supervision
MCV:  data curation
DNK:  data curation
AU: methodology
FI: writing review \& editing, supervision
AAH: writing review \& editing, supervision 
FWA: conceptualization, writing review \& editing, supervision
IC: conceptualization, methodology, writing review \& editing, supervision

\section*{Acknowledgments}
The present research has been partially supported by MUR, grant Dipartimento di Eccellenza 2023-2027 awarded to Department of Mathematics, Politecnico di Milano. This work received funding from the European Union’s Horizon Europe research and innovation program under Grant Agreement No. 101057849 (DataTools4Heart project) and No. 101080430 (AI4HF project). This publication is part of the project “Computational medicine for cardiac disease” with file number 2023.022 of the research programme “Computing Time on National Computer Facilities” which is (partly) financed by the Dutch Research Council (NWO).
FWA is supported by UCL Hospitals NIHR Biomedical Research Centre.
IC is funded by the project CaRe-NLP with file number NGF.1607.22.014 of the research programme AiNed Fellowship Grants which is (partly) financed by the Dutch Research Council (NWO).

%TC:ignore

%\bibliographystyle{plain}
\bibliographystyle{unsrt}
\bibliography{ref}

%USE THE BELOW OPTIONS IN CASE YOU NEED AUTHOR YEAR FORMAT.
%\bibliographystyle{abbrvnat}
%\bibliography{reference}

% %% sample for biography with author's image
% \begin{biography}{{\color{black!20}\rule{77pt}{77pt}}}{\author{Author Name.} This is sample author biography text. The values provided in the optional argument are meant for sample purposes. There is no need to include the width and height of an image in the optional argument for live articles. This is sample author biography text this is sample author biography text this is sample author biography text this is sample author biography text this is sample author biography text this is sample author biography text this is sample author biography text this is sample author biography text.}
% \end{biography}
% %% sample for biography without author's image
% \begin{biography}{}{\author{Author Name.} This is sample author biography text this is sample author biography text this is sample author biography text this is sample author biography text this is sample author biography text this is sample author biography text this is sample author biography text this is sample author biography text.}
% \end{biography}
%TC:endignore

\clearpage

\begin{appendices}
\renewcommand\thetable{\thesection.\arabic{table}}
\renewcommand\thefigure{\thesection.\arabic{figure}}  

\definecolor{lightgreen}{HTML}{02f206} % Hex color definition
\definecolor{darkgreen}{HTML}{016904}

\newcommand{\hlred}[1]{\sethlcolor{red}\hl{#1}}
\newcommand{\hldarkgreen}[1]{\sethlcolor{darkgreen}\hl{#1}}
\newcommand{\hllightgreen}[1]{\sethlcolor{lightgreen}\hl{#1}}

\pagenumbering{Roman}

\section{Appendix A - Population characteristics}
\setcounter{table}{0}
\setcounter{figure}{0}  
\label{app:A}
\subsection{ICD-10-CM codes for cohort selection of hospitalized HF patients}
The following ICD-10-CM codes are those used to select the hospitalizations to be included in our cohort:
\begin{itemize}
    \item	I50 Heart failure
    \item	I11 Hypertensive heart disease
    \item	I13.0 Hypertensive heart and chronic kidney disease with heart failure and stage 1 through stage 4 chronic kidney disease, or unspecified chronic kidney disease
    \item	I13.2 Hypertensive heart and chronic kidney disease with heart failure and with stage 5 chronic kidney disease, or end stage renal disease
    \item	I26.0 Pulmonary embolism with acute cor pulmonale
    \item	I09.81 Rheumatic heart failure
    \item	I97.13 Postprocedural heart failure
\end{itemize}

Table \ref{tab:population} summarizes the characteristics of the population with respect to the structured covariates.
\begin{table*}[p]
\scriptsize
\caption{Covariates for the models based on structured data with their distribution and missing values.  Comorbidities and medication features do not have missing values since the absence of a code in the dataset is assumed to be equal to a negative value.}
\label{tab:population}
\centering
\begin{tabular}{@{}lllr@{}}
\toprule
\textbf{Feature}                           & \textbf{Values} & \textbf{N (\%)}    & \textbf{Missing   values (\%)} \\ \midrule
\multirow{2}{*}{Age}                   & \textless{}= 75    & 20,730   (62.6\%)  & \multirow{2}{*}{0.0 \%}    \\ 
                                       & \textgreater 75    & 12,375   (37.4\%)  &                          \\ \midrule
\multirow{2}{*}{Gender}                & Male               & 19,386 (58.6   \%) & \multirow{2}{*}{0.0 \%}    \\ 
                                       & Female             & 13,719 (41.4   \%) &                          \\ \midrule
\multirow{2}{*}{MAP}                   & \textless 90.0       & 11,774 (35.6   \%) & \multirow{2}{*}{3.5 \%}  \\ 
                                       & \textgreater{}= 90.0 & 21,331 (64.4 \%)   &                          \\ \midrule
\multirow{2}{*}{Heart Rate}                & \textless 70.0    & 9,888 (29.9   \%)  & \multirow{2}{*}{3.6 \%}           \\ 
                                       & \textgreater{}= 70.0 & 23,217 (70.0.1   \%) &                          \\ \midrule
\multirow{4}{*}{BMI}                   & \textless{}= 18.5  & 1,100 (3.3 \%)     & \multirow{4}{*}{7.5 \%}  \\ 
                                       & (18.5, 25{]}       & 10,983 (33.2   \%) &                          \\ 
                                       & (25,30.0)            & 12,280 (37.1   \%) &                          \\ 
                                       & \textgreater{}= 30.0 & 8,742 (26.4   \%)  &                          \\ \midrule
\multirow{4}{*}{EGFR}                  & \textgreater{}= 90.0 & 2,801 (8.5 \%)     &                          \\ 
                                       & (60.0,90.0)            & 13,382   (40.0.4\%)  & \multirow{3}{*}{27 \%}   \\ 
                                       & (30.0,60.0{]}          & 13,765 (41.6   \%) &                          \\ 
                                       & \textless{}= 30.0    & 3,157 (9.5 \%)     &                          \\ \midrule
\multirow{2}{*}{Ischaemic   heart disease} & True            & 11,530 (34.8   \%) & \multirow{2}{*}{0.0 \%}             \\ 
                                       & False              & 21,575 (65.2   \%) &                          \\ \midrule
\multirow{2}{*}{Anaemia}               & True               & 4,759 (14.4   \%)  & \multirow{2}{*}{0.0 \%}    \\ 
                                       & False              & 28,346 (85.6   \%) &                          \\ \midrule
\multirow{2}{*}{Atrial   Fibrillation} & True               & 9,550 (28.8\%)     & \multirow{2}{*}{0.0 \%}    \\ 
                                       & False              & 23,555 (71.2   \%) &                          \\ \midrule
\multirow{2}{*}{Diabetes}              & True               & 9,008 (27.2   \%)  & \multirow{2}{*}{0.0 \%}    \\ 
                                       & False              & 24,097 (72.8   \%) &                          \\ \midrule
\multirow{2}{*}{Hypertension}          & True               & 12,766 (38.6   \%) & \multirow{2}{*}{0.0 \%}    \\ 
                                       & False              & 20,339 (61.4   \%) &                          \\ \midrule
\multirow{2}{*}{COPD}                  & True               & 4,587 (13.8   \%)  & \multirow{2}{*}{0.0 \%}    \\ 
                                       & False              & 28,527 (86.2   \%) &                          \\ \midrule
\multirow{2}{*}{Valvular   Disease}    & True               & 6,241 (18.9   \%)  & \multirow{2}{*}{0.0 \%}    \\ 
                                       & False              & 26,864 (81.1   \%) &                          \\ \midrule
\multirow{2}{*}{Cancer in past   3 years}  & True            & 7,782 (23.5   \%)  & \multirow{2}{*}{0.0 \%}             \\ 
                                       & False              & 25,323 (76.5   \%) &                          \\ \midrule
\multirow{2}{*}{Device   therapy}      & True               & 4,900 (14.8   \%)  & \multirow{2}{*}{0.0 \%}    \\ 
                                       & False              & 28,205 (85.2   \%) &                          \\ \midrule
\multirow{2}{*}{RASi}                  & True               & 16,540 (50.0   \%) & \multirow{2}{*}{0.0 \%}    \\ 
                                       & False              & 16,565 (50.0   \%) &                          \\ \midrule
\multirow{2}{*}{Beta Blockers}         & True               & 20,636 (62.3   \%) & \multirow{2}{*}{0.0 \%}    \\ 
                                       & False              & 12,469   (37.7\%)  &                          \\ \midrule
\multirow{2}{*}{MRA}                   & True               & 10,807   (32.6\%)  & \multirow{2}{*}{0.0 \%}    \\ 
                                       & False              & 22,298 (67.4   \%) &                          \\ \midrule
\multirow{2}{*}{Digoxin}               & True               & 4,617 (13.9   \%)  & \multirow{2}{*}{0.0 \%}    \\ 
                                       & False              & 28,488 (86.1   \%) &                          \\ \midrule
\multirow{2}{*}{LoopDiuretics}         & True               & 20,836 (62.9   \%) & \multirow{2}{*}{0.0 \%}    \\ 
                                       & False              & 12,269 (37.1   \%) &                         \\ \bottomrule
\end{tabular}
\end{table*}

\subsection{Differences between AMC and VuMC populations}
The dataset includes data from both locations of Amsterdam UMC: AMC and VUmc, two separate hospitals located in Amsterdam. In each hospital, the entire population of hospitalized HF patients during the selected time period was considered. To assess whether there were significant differences between the two populations, we trained a logistic regression model to classify between AMC and VUmc using the structured covariates. This model achieved an AUC of 0.5910 (standard deviation 0.0080), which does not indicate strong differences between the two populations.

\section{Appendix B - Silver and gold labelling}
\setcounter{table}{0}    
\label{app:B}
\subsection{Silver labelling}
\label{app:silver}
In this section, we detail the lists of ICD-10-CM and SNOMED-CT codes that, when present in the \textit{diagnosis}, \textit{past history} or \textit{problem list} tables, allow us to derive a HFrEF (systolic) or HFpEF (diastolic) silver label, and we provide some details about the LVEF values estimation from echocardiographic reports and text mentions.

\subsubsection{ICD-10-CM codes specifying systolic/diastolic HF}
\begin{itemize}
    \item	I50.20 Unspecified systolic (congestive) heart failure
\item	I50.21 Acute systolic (congestive) heart failure
\item	I50.22 Chronic systolic (congestive) heart failure
\item	I50.23 Acute on chronic systolic (congestive) heart failure
\item	I50.30 Unspecified diastolic (congestive) heart failure
\item	I50.31 Acute diastolic (congestive) heart failure
\item	I50.32 Chronic diastolic (congestive) heart failure
\item	I50.33 Acute on chronic diastolic (congestive) heart failure

\end{itemize}

\subsubsection{SNOMED-CT codes specifying systolic/diastolic HF}
\begin{itemize}
    \item 417996009 Systolic heart failure (disorder)
\item	418304008 Diastolic heart failure (disorder)
\item	426263006 Congestive heart failure due to left ventricular systolic dysfunction (disorder)
\item	441481004 Chronic systolic heart failure
\item	441530006 Chronic diastolic heart failure
\item	443254009 Acute systolic heart failure (disorder)
\item	443253003 Acute on chronic systolic heart failure (disorder)
\item	443343001 Acute diastolic heart failure (disorder)
\item	443344007 Acute on chronic diastolic heart failure (disorder)
\item	120851000119104 Systolic heart failure stage D
\item	120861000119102 Systolic heart failure stage C
\item	120871000119108 Systolic heart failure stage B
\item	120881000119106 Diastolic heart failure stage D
\item	120891000119109 Diastolic heart failure stage C
\item	120901000119108 Diastolic heart failure stage B
\item	15629641000119107 Systolic heart failure stage B due to ischaemic cardiomyopathy (disorder)
\item	15629741000119102 Systolic heart failure stage C due to ischaemic cardiomyopathy (disorder)
\end{itemize}

\subsubsection{LVEF estimation from echocardiographies}
LVEF can be estimated/measured from echocardiographic images using different techniques, some of which are more reliable than others. Because of this, we define a priority order to be used in case multiple values, estimated/measured with different techniques, are available for the same patient. From the most reliable to the least reliable:
\begin{enumerate}
    \item 4D and 3D estimation methods
    \item Biplane measurements, including automatic calculations and  manual calculations using both Apical 2 Chamber (A2C) and Apical 4 Chamber (A4C) views
    \item Single-plane measurements, including: automatic and manual calculations from A2C or A4C views; area-length method; cube formula; geometric modelling
    \item Teichholz estimation method
\end{enumerate}

In some cases, a range of estimated LVEF values is reported in echocardiographic results. In these cases, we discard results with range $> 10\%$, since they are not reliable and they are likely to indicate issues in the image acquisition. For those with range $\leq 10 \%$, we consider the lower bound of the range.

\subsubsection{LVEF extraction from text}
To extract explicit mention of LVEF from the text of discharge letters, we use the following regular expression:
\begin{lstlisting}[breaklines]
(?:ejection fraction|ejectiefractie|(lv)?ef):?\s*((?:100|\d{1,2})(?:\.\d+)?(?:\s*-\s*(?:100|\d{1,2})(?:\.\d+))?)
\end{lstlisting}
which captures integer or decimal numbers, possibly with ranges. We also check for explicit mentions of \textit{systolic dysfunction} or \textit{diastolic dysfunction}, not preceded by negation. 

\subsection{Gold labelling}
\label{app:gold}
Hospitalizations for 300 patients were manually labelled by MB. For 131 patients it was not possible to assign an HFpEF or HFrEF label with complete certainty, so they were later excluded from the evaluation. Of these, only 1 was certainly belonging to the HFmrEF class. 

Table \ref{tab:gold_labels_ct} reports the distribution of the manually annotated gold labels, compared also with the corresponding silver labels. Considering only patients with a specified silver and gold label, this leads to Cohen's Kappa of 0.383 and a Krippendorff's Alpha of 0.378 between silver and gold labels. Given this, the discrepancy between performance on the gold-labelled and external validation sets can be probably attributed mostly to the limited sample size and to larger class unbalance present in the gold-labelled set.

\begin{table}[ht]
\caption{Contingency table of gold vs silver labels on the gold-labelled dataset}
\label{tab:gold_labels_ct}
\centering
\begin{tabular}{lccc}
\toprule
\textbf{Gold \textbackslash Silver label} & \textbf{HFpEF} & \textbf{HFrEF} & \textbf{Unspecified} \\
\midrule
\textbf{HFpEF} & 4 & 2 & 15 \\
\textbf{HFrEF} & 8 & 66 & 74 \\
\textbf{Unspecified} & 8 & 16 & 107 \\
\bottomrule
\end{tabular}
\end{table}

\section{Appendix C - Classification models}
\setcounter{table}{0}    
\label{app:C}
In this section, we provide additional details on our classification models, in addition to the information provided in the Material and Methods section of the paper.

\subsection{Classification from structured data}
Numerical features are standardized and missing values are imputed using the \textit{IterativeImputer} method of \textit{scikit-learn} Python library.

LR models are regularized with L2 regularization, selecting the regularization coefficient with grid search (Table \ref{tab:l2-reg}). 

For EBM models, the learning rate was selected via grid search (Table \ref{tab:ebm-lr}).

\begin{table}[hp]
\caption{10-fold cross-validation classification results on the training dataset of the LR model on structured data with different values of the regularization parameter $C = 1 / \lambda$.
%P = precision. R = recall. F1 = F1-score. AUC = Area under the receiver operating characteristic curve.
}
\label{tab:l2-reg}
\centering
\begin{tabular}{ccccccc}
\toprule
L2 Reg (C) & \makecell{P {[}\%{]} \\ (std)} & \makecell{R {[}\%{]} \\ (std)} & \makecell{F1 {[}\%{]} \\ (std)} & \makecell{AUC {[}\%{]} \\ (std)} \\ 
\midrule

$1 \cdot 10^{-3}$ & \makecell{68.59 \\ (1.70)} & \makecell{66.11 \\ (1.40)} & \makecell{66.15 \\ (1.50)} & \makecell{76.35 \\ (1.40)} \\
$1 \cdot 10^{-2}$ & \makecell{68.59 \\ (1.70)} & \makecell{66.11 \\ (14.00)} & \makecell{66.15 \\ (1.50)} & \makecell{76.35 \\ (1.40)} \\
$1 \cdot 10^{-1}$ & \makecell{68.80 \\ (1.20)} & \makecell{\textbf{68.48} \\ (1.00)} & \makecell{\textbf{68.64} \\ (1.00)} & \makecell{76.42 \\ (1.40)} \\
$1 \cdot 10^{0}$& \makecell{\textbf{68.91} \\ (1.30)} & \makecell{\textbf{68.48} \\ (1.10)} & \makecell{68.58 \\ (1.10)} & \makecell{\textbf{76.48} \\ (1.40)} \\
$1 \cdot 10^{1}$ & \makecell{\textbf{68.91} \\ (1.30)} & \makecell{\textbf{68.48} \\ (1.10)} & \makecell{68.58 \\ (1.10)} & \makecell{\textbf{76.48} \\ (1.40)} \\
$1 \cdot 10^{2}$ & \makecell{\textbf{68.91} \\ (1.30)} & \makecell{\textbf{68.48} \\ (1.10)} & \makecell{68.58 \\ (1.10)} & \makecell{\textbf{76.48} \\ (1.40)} \\
$1 \cdot 10^{3}$ & \makecell{\textbf{68.91} \\ (1.30)} & \makecell{\textbf{68.48} \\ (1.10)} & \makecell{68.58 \\ (1.10)} & \makecell{\textbf{76.48} \\ (1.40)} \\
no reg & \makecell{\textbf{68.91} \\ (1.30)} & \makecell{\textbf{68.48} \\ (1.10)} & \makecell{68.58 \\ (1.10)} & \makecell{\textbf{76.48} \\ (1.40)} \\
\bottomrule
\end{tabular}
\end{table}

\begin{table}[hp]
\caption{10-fold CV results on training data for explainable boosting machine models on structured data, with different learning rates.}
\label{tab:ebm-lr}
\centering
\begin{tabular}{ccccc}
\toprule
Learning Rate & \makecell{P {[}\%{]} \\ (std)} & \makecell{R {[}\%{]} \\ (std)} & \makecell{F1 {[}\%{]} \\ (std)} & \makecell{AUC {[}\%{]} \\ (std)} \\ 
\midrule

$2 \cdot 10^{-2}$ & \makecell{74.42 \\ (1.10)}  & \makecell{\textbf{70.45} \\ (1.00)}  & \makecell{72.38 \\ (1.00)}  & \makecell{77.40 \\ (1.40)} \\
$5 \cdot 10^{-2}$ & \makecell{\textbf{74.60} \\ (1.10)}  & \makecell{70.40 \\ (1.00)}  & \makecell{\textbf{72.80} \\ (1.00)}  & \makecell{\textbf{77.45} \\ (1.40)} \\
$2 \cdot 10^{-3}$ & \makecell{74.42 \\ (1.10)}  & \makecell{\textbf{70.45} \\ (1.00)}  & \makecell{72.38 \\ (1.00)}  & \makecell{77.40 \\ (1.40)} \\
${5 \cdot 10^{-3}}$ & \makecell{74.42 \\ (1.10)}  & \makecell{\textbf{70.45} \\ (1.00)}  & \makecell{72.38 \\ (1.00)}  & \makecell{77.40 \\ (1.40)} \\
\bottomrule
\end{tabular}
\end{table}

\subsection{Classification from discharge letters}
\label{app:classification_from_discharge_letters}
\subsubsection{Training settings and hyperparameters}
For BERT-based models, we experimented with fine-tuning only the last and only the last three layers. 

We compared results by keeping only the first 512 tokens of the discharge letters and dividing the letters into 512 tokens-long chunks that are processed in parallel, taking their maximum probability output at the end.

We also compared the effect of LVEF masking on training data only, test data only or both.
Results of these experiments are summarized in Table \ref{tab:hyperparam-bb}.

\begin{table}[htb]
\centering
\caption{Classification results on training data for \geitje{} with different values for the temperature parameter}
\label{tab:temp-geitje}
\begin{tabular}{@{}cccc@{}}
\toprule
Temperature & P [\%] & R [\%] & F1 [\%] \\
\midrule
0.1                  & 77.21              & 75.63           & 76.41       \\
0.2                  & \textbf{78.10}              & \textbf{76.42}           & \textbf{77.38}       \\
0.3                  & 75.21              & 73.52           & 74.36       \\
0.4                  & 72.31              & 69.55           & 70.90 \\
\bottomrule
\end{tabular}
\end{table}

\begin{table}[hp]
\centering
\small
\caption{10-fold cross-validation classification results on the training dataset for black-box models with different numbers of fine-tuned layers, different masking of ejection fraction and with/without truncation to 512 tokens. 
%P = precision. R = recall. F1 = F1-score. AUC = Area under the receiver operating characteristic curve.
}
\label{tab:hyperparam-bb}
\vspace{0.5em}
\begin{tabular}{lcccc}
\toprule
Model & \makecell{P {[}\%{]} \\ (std)} & \makecell{R {[}\%{]} \\ (std)} & \makecell{F1 {[}\%{]} \\ (std)} & \makecell{AUC {[}\%{]} \\ (std)} \\ 
\midrule
\multicolumn{5}{c}{Truncation to 512 Tokens - EF Always Masked} \\ 
\midrule
\medrobertanl{} 1 layer FT & \makecell{84.35 \\ (2.0)} & \makecell{53.17 \\ (10.5)} & \makecell{64.53 \\ (8.0)} & \makecell{62.17 \\ (1.1)} \\
\medrobertanl{} 3 layers FT & \makecell{\textbf{84.53} \\ (2.9)} & \makecell{\textbf{55.64} \\ (10.2)} & \makecell{\textbf{66.51} \\ (7.29)} & \makecell{\textbf{63.71} \\ (3.0)} \\
\robbert{} 1 layer FT & \makecell{80.59 \\ (1.9)} & \makecell{45.38 \\ (11.5)} & \makecell{57.27 \\ (9.2)} & \makecell{55.53 \\ (2.0)} \\
\robbert{} 3 layers FT & \makecell{82.12 \\ (3.2)} & \makecell{50.49 \\ (19.1)} & \makecell{60.05 \\ (14.7)} & \makecell{57.47 \\ (2.0)} \\
\geitje{} & \makecell{78.20 \\ (2.1)} & \makecell{42.74 \\ (3.3)} & \makecell{55.22 \\ (3.0)} & / \\
\midrule
\multicolumn{5}{c}{Truncation to 512 Tokens - EF Masked Only in Training} \\
\midrule
\medrobertanl{} 1 layer FT & \makecell{83.42 \\ (2.1)} & \makecell{55.07 \\ (7.2)} & \makecell{65.98 \\ (4.8)} & \makecell{61.93 \\ (2.5)} \\
\medrobertanl{} 3 layers FT & \makecell{\textbf{84.52} \\ (2.5)} & \makecell{\textbf{55.56} \\ (7.5)} & \makecell{\textbf{66.61} \\ (5.1)} & \makecell{\textbf{63.40} \\ (2.7)} \\
\robbert{} 1 layer FT & \makecell{80.51 \\ (2.2)} & \makecell{46.15 \\ (7.5)} & \makecell{58.35 \\ (6.3)} & \makecell{55.35 \\ (2.8)} \\
\robbert{} 3 layers FT & \makecell{83.45 \\ (3.2)} & \makecell{46.23 \\ (20.1)} & \makecell{51.16 \\ (19.2)} & \makecell{58.90 \\ (2.2)} \\
\geitje{} & / & / & / & / \\
\midrule
\multicolumn{5}{c}{Truncation to 512 Tokens - EF Never Masked} \\
\midrule
\medrobertanl{} 1 layer FT & \makecell{83.44 \\ (1.8)} & \makecell{53.39 \\ (3.8)} & \makecell{65.03 \\ (2.8)} & \makecell{61.87 \\ (2.9)} \\
\medrobertanl{} 3 layers FT & \makecell{\textbf{83.98} \\ (2.5)} & \makecell{\textbf{58.57} \\ (8.6)} & \makecell{\textbf{68.53} \\ (5.8)} & \makecell{\textbf{63.28} \\ (1.7)} \\
\robbert{} 1 layer FT & \makecell{81.13 \\ (2.2)} & \makecell{47.01 \\ (9.7)} & \makecell{58.94 \\ (7.0)} & \makecell{55.39 \\ (3.1)} \\
\robbert{} 3 layers FT & \makecell{81.61 \\ (2.1)} & \makecell{52.77 \\ (12.4)} & \makecell{63.20 \\ (9.7)} & \makecell{58.15 \\ (2.2)} \\
\geitje{} & \makecell{78.20 \\ (2.0)} & \makecell{42.74 \\ (3.3)} & \makecell{55.22 \\ (3.0)} & / \\
\midrule
\multicolumn{5}{c}{Chunking with Max Prob - EF Masked Only in Training} \\
\midrule
\medrobertanl{} 1 layer FT & \makecell{87.95 \\ (1.0)} & \makecell{74.64 \\ (6.4)} & \makecell{80.56 \\ (3.6)} & \makecell{81.77 \\ (0.6)} \\
\medrobertanl{} 3 layers FT & \makecell{\textbf{88.50} \\ (1.1)} & \makecell{75.10 \\ (4.5)} & \makecell{81.45 \\ (2.1)} & \makecell{\textbf{85.03} \\ (0.8)} \\
\robbert{} 1 layer FT & \makecell{83.91 \\ (0.9)} & \makecell{82.56 \\ (5.2)} & \makecell{\textbf{82.64} \\ (2.3)} & \makecell{75.11 \\ (0.5)} \\
\robbert{} 3 layers FT & \makecell{83.50 \\ (2.1)} & \makecell{\textbf{84.55} \\ (3.4)} & \makecell{82.40 \\ (3.1)} & \makecell{73.49 \\ (1.3)} \\
\geitje{} & \makecell{78.10 \\ (4.2)} & \makecell{76.42 \\ (2.4)} & \makecell{77.38 \\ (5.0)} & / \\
\bottomrule
\end{tabular}
\end{table}  

For the \geitje{} model, the temperature parameter was selected with grid search (Table \ref{tab:temp-geitje}). For TF-IDF baselines, L1 regularization was employed, removing punctuation and stop-words.

\subsubsection{MedRoberta.nl potential overlapping in pre-training set}
Since the MedRoberta.nl model was pre-trained by its authors on a dataset of 12.3 GB of clinical notes from Amsterdam UMC, this might partially overlap with our dataset. In particular, they used data from 2017 and 2020 for VUmc location. Because of this, we compared performances on our entire VUmc dataset with those on our entire VUmc dataset without these two years and with those on our VUmc dataset with only these two years. Results, reported in Table~\ref{tab:app-vumc-overlap}, confirm the absence of a significative difference.

\begin{table*}
\centering
\caption{Results on VUmc dataset stratifying by group of years that might (2017,2020) or might not overlap with the pre-training dataset of MedRoberta.nl. 
%P = precision. R = recall. F1 = F1-score. AUC = Area under the receiver operating characteristic curve.
}
\label{tab:app-vumc-overlap}
\begin{tabular}{@{}llrrrrr@{}}
\toprule
Model         & Dataset                 & Dataset size & P [\%] & R [\%] & F1 [\%] & AUC [\%]  \\ \midrule
\auggamlr{}    & All VUmc                & 1098 (100\%) & 74.01     & 73.36  & 73.68    & 80.77 \\
\auggamlr{}    & VUmc w/o 2017 and 2020  & 795 (77\%)   & 76.61     & 75.90  & 76.26    & 81.55 \\
\auggamlr{}    & VUmc only 2017 and 2020 & 303 (23\%)   & 67.94     & 67.21  & 67.56    & 75.01 \\ \midrule
\auggamebm{}   & All VUmc                & 1098 (100\%) & 75.27     & 79.97  & 75.12    & 80.10 \\
\auggamebm{}    & VUmc w/o 2017 and 2020  & 795 (77\%)   & 73.00     & 80.44  & 76.54    & 80.77 \\
\auggamebm{}    & VUmc only 2017 and 2020 & 303 (23\%)   & 70.73     & 79.35  & 74.79    & 79.82 \\ \midrule
\medrobertanl{} & All VUmc                & 1098 (100\%) & 84.44     & 74.98  & 80.15    & 83.52 \\
\medrobertanl{} & VUmc w/o 2017 and 2020  & 795 (77\%)   & 85.33     & 75.41  & 81.77    & 83.85 \\
\medrobertanl{} & VUmc only 2017 and 2020 & 303 (23\%)   & 82.15     & 73.12  & 79.52    & 83.22 \\ \bottomrule
\end{tabular}
\end{table*}

\subsection{Classification from structured data and discharge letters}
For models using both structured and unstructured data, structured data were pre-processed in the same way as for models with structured data only, for what concerns missing values and standardization.

\subsection{Additional classification results}
Table~\ref{tab:res-train} reports classification results on the silver labelled training data from AMC hospital, while Table~\ref{tab:ext_val_per_class} reports result son the external validation set separated per class.

\begin{table*}[tbph]
\centering
\caption{10-fold cross-validation classification results on the training dataset. We show results for models that use structured data only, discharge notes only (baselines using TF-IDF representations, black-box, and white-box models, respectively), and that combine structured and unstructured data. 
%P = precision. R = recall. F1 = F1-score. AUC = Area under the receiver operating characteristic curve.
}
\label{tab:res-train}
\vspace{0.5em}
\begin{tabular}{@{}lr@{\hskip 0.2in}rrrr@{}}
\toprule
& Model & \makecell{P {[}\%{]} \\ (std)} & \makecell{R {[}\%{]} \\ (std)} & \makecell{F1 {[}\%{]} \\ (std)} & \makecell{AUC {[}\%{]} \\ (std)} \\ 
\midrule
\multirow{3}{*}{\rotatebox[origin=c]{90}{\makecell{Struct. \\ data}}} 
& \uijletalorig{} & \makecell{66.56} & \makecell{66.50} & \makecell{66.55} & \makecell{69.76}  \\
& \uijletalstruct{}  & \makecell{68.80 \\ (1.2)} & \makecell{68.48 \\ (1.0)} & \makecell{68.64 \\ (1.0)} & \makecell{76.42 \\ (1.4)}  \\
& \ebmstruct{}       & \makecell{74.42 \\ (1.1)} & \makecell{70.45 \\ (1.0)} & \makecell{72.38 \\ (1.0)} & \makecell{77.40 \\ (1.4)}  \\ 
\midrule
\multirow{3}{*}{\rotatebox[origin=c]{90}{\makecell{Unstructured \\ data (discharge \\ letters)}}} 
& \lrtfidf{}         & \makecell{64.40 \\ (1.1)} & \makecell{73.61 \\ (0.9)} & \makecell{68.69 \\ (1.0)} & \makecell{76.10 \\ (1.2)}  \\
& \ebmtfidf{}        & \makecell{68.62 \\ (1.2)} & \makecell{71.56 \\ (1.2)} & \makecell{70.06 \\ (1.2)} & \makecell{75.28 \\ (1.3)}  \\
\cmidrule{3-6}
& \medrobertanl{}    & \makecell{\textbf{88.50} \\ (1.1)} & \makecell{75.10 \\ (4.5)} & \makecell{\textbf{81.45} \\ (2.1)} & \makecell{\textbf{85.03} \\ (0.8)}  \\
& \robbert{}         & \makecell{79.50 \\ (2.4)} & \makecell{\textbf{84.55} \\ (3.4)} & \makecell{80.37 \\ (3.1)} & \makecell{73.49 \\ (1.3)}  \\
& \geitje{}          & \makecell{78.10 \\ (4.2)} & \makecell{76.42 \\ (2.4)} & \makecell{77.38 \\ (5.0)} & -      \\
\cmidrule{3-6}
& \auggamlr{}        & \makecell{71.65 \\ (1.0)} & \makecell{74.84 \\ (1.1)} & \makecell{73.56 \\ (1.0)} & \makecell{85.12 \\ (0.9)}  \\
& \auggamebm{}       & \makecell{70.04 \\ (1.0)} & \makecell{73.21 \\ (0.9)} & \makecell{71.10 \\ (1.0)} & \makecell{83.42 \\ (1.2)}  \\ 
\midrule
\multirow{2}{*}{\rotatebox[origin=c]{90}{\makecell{Both}}} 
& \auggamlrstruct{}  & \makecell{72.22 \\ (1.3)} & \makecell{73.55 \\ (1.1)} & \makecell{72.84 \\ (1.2)} & \makecell{84.54 \\ (1.2)}  \\
& \auggamebmstruct{} & \makecell{73.74 \\ (1.1)} & \makecell{76.88 \\ (1.0)} & \makecell{74.86 \\ (1.1)} & \makecell{84.83 \\ (1.1)}  \\ 
\bottomrule
\end{tabular}

\end{table*}

\begin{table*}[]
\centering
\small
\caption{External validation results separated for HFrEF and HFpEF}
\label{tab:ext_val_per_class}
\begin{tabular}{@{}lr@{\hskip 0.2in}rrrr@{\hskip 0.2in}rrrr@{}}
\toprule
&& \multicolumn{4}{c}{HFrEF} & \multicolumn{4}{c}{HFpEF} \\ %\midrule
&  Model & P {[}\%{]} & R {[}\%{]} & F1 {[}\%{]} & AUC {[}\%{]} & P {[}\%{]} & R {[}\%{]} & F1 {[}\%{]} & AUC {[}\%{]} \\ \midrule
\multirow{3}{*}{\rotatebox[origin=c]{90}{\parbox[c]{1cm}{\centering Struct. data}}}
  & \uijletalorig{40} &
  71.80 &
  70.73 &
  71.26 &
  73.70 &
  38.24 &
  38.17 &
  38.20 &
  50.46 \\
                        & \uijletalorig{50} & 72.11 & 71.22 & 71.66 & 74.56 & 37.31 & 36.70 & 37.00 & 47.88 \\
                        & \uijletalstruct & 70.47 & 68.89 & 69.67 & 78.90 & 54.35 & 56.45 & 55.38 & 59.50 \\
                        & \ebmstruct      & 76.91 & 74.56 & 75.72 & 78.50 & 63.51 & 52.80 & 57.66 & 67.14 \\ \midrule
\multirow{3}{*}{\rotatebox[origin=c]{90}{\parbox[c]{3cm}{\centering Unstructured data (discharge letters)}}}
  & \lrtfidf &
  67.58 &
  74.56 &
  70.90 &
  78.30 &
  44.38 &
  64.24 &
  52.49 &
  61.18 \\
                        & \ebmtfidf       & 65.47 & 70.87 & 68.06 & 74.09 & 57.67 & 61.15 & 59.36 & 63.01 \\ \cmidrule{3-10}
                        & \medrobertanl   & \textbf{89.51} & 78.55 & \textbf{83.67} & \textbf{86.88} & 69.23 & 64.27 & 66.66 & 73.44 \\
                        & \robbert        & 84.61 & 81.24 & 82.89 & 83.11 & \textbf{70.97} & \textbf{72.23} & \textbf{71.60} & \textbf{73.87} \\
                        & \geitje         & 81.14 & 78.17 & 79.54 & -     & 62.62 & 55.13 & 58.64 & -     \\ \cmidrule{3-10}
                        & \auggamlr       & 78.45 & 75.67 & 77.03 & 83.44 & 60.69 & 66.43 & 63.43 & 72.76 \\
                        & \auggamebm      & 75.88 & \textbf{84.23} & 78.46 & 83.45 & 62.64 & 67.19 & 64.84 & 70.05 \\ \midrule
\multirow{2}{*}{\rotatebox[origin=c]{90}{\parbox[c]{0.9cm}{\centering Both}}}
  & \auggamlrstruct &
  76.13 &
  76.78 &
  76.45 &
  82.32 &
  64.09 &
  59.46 &
  61.69 &
  73.52 \\
                        & \auggamlrstruct & 73.49 & 75.67 & 74.56 & 81.21 & 63.93 & 63.15 & 65.54 & 73.77 \\
\bottomrule   
\end{tabular}
\end{table*}

\clearpage
\section{Appendix D - Explainability techniques}
\label{app:D}
\setcounter{table}{0}    
In this section, we provide additional details on the explainability methods. For LIME and SHAP, we refer both to their original papers and the specific implementation details of the \texttt{lime} and \texttt{shap} Python packages.
\subsection{LIME}
\label{app:lime}
For a given point, LIME builds a linear model by sampling n~(=100) perturbed version of that point and using this set of points to train the linear model. The linear model is trained by weighting these samples with weights that are inversely proportional to their distance from the original point to be explained. The weights of the features in this linear model become the feature importance scores for that point.
    
The global explanations can be derived by averaging the feature importance scores on m (=100) points.
    
In particular, for textual data, the sampling of the perturbed points is obtained in the following way, for each document $x$ to be explained, for each perturbed instance $x_i$ to be created:
    \begin{enumerate}
        \item Randomly draw $s_i$ in $[1,d]$, where $d$ is the number of distinct words in $x$
        \item Randomly draw a subset $S_i 	\subseteq \{1,..,d\}$ with cardinality $s_i$
        \item All the words in $x$ with indices in $S_i$ are removed from $x$, generating $x_i$
        \item Define $z_i \in \{0,1\}^d$ as a binary vector representing the absence or presence of the original words of $x$ in $x_i$
        \item The weight of $z_i$ in the linear model is defined as $$\pi_i = \sqrt(exp(\frac{-(cos\_dist(\textbf{1}, z_i) \cdot 100)^2}{\nu^2}))$$
        with $\nu=25$ 
        In this way, the weight depends only on the number of deleted words
    \end{enumerate}
    The linear model is a weighted ridge regression fitted on $z_1, ..., z_n$ with the weights $\pi_1, ..., \pi_n$ and regularization parameter $\lambda=1$. The labels of the perturbed points are obtained by applying the classification model to be explained. Before fitting this model, a feature selection mechanism is applied. Forward selection is used if the number of features is $\leq6$; otherwise, the top K (=10) features with the highest absolute weights in the model fitted with all the features (with $\lambda=0.01)$ are selected.

\subsection{SHAP}
\label{app:shap}
The exact computation of Shapley values, as defined in Game Theory, would require, for a given document, to compute the model output, for each token $t_i$ in the document, on all the possible documents that can be created by removing that token and possibly other ones. 

In particular, for each of these subsets of remaining tokens S, one should compute $f(S \cup {t_i}) - f(S)$ and then compute a weighted sum of these results to get the Shapley value for $t_i$. SHAP approximates this computation. There are multiple methods that can be adopted, but for Transformers-based models, the suggested method is the \textit{partition explainer}, which computes the so-called Owen values. With this method, features (tokens) are grouped into coalitions. To calculate the contribution of each token, the weighted sum is over all the coalitions which do not contain that token and on all the other tokens in its coalition. The coalitions are built by applying an ad-hoc hierarchical agglomerative clustering over the tokens. At each step, it scores a pair of consecutive coalitions with a heuristic function based on punctuation signs and connectors that try to preserve the sentence structure. 
Global explanations are derived, as with LIME, by averaging over feature scores on m (=100) samples.

\subsection{Aug-Linear}
\label{app:auggam}
Aug-Linear models are composed of two steps:
\begin{enumerate}
    \item Extraction of $n$-grams
    \item Embedding of $n$-grams
\end{enumerate}

For $n$-grams extraction, we experiment with $n$ from 1 to 5, and at each $n$, we filter out n-grams with a frequency lower than a threshold, selected via grid search (see Tables \ref{tab:auggam-lr-thresholds} and \ref{tab:auggam-ebm-thresholds}).

\begin{table}[hp]
\centering
\caption{10-fold cross-validation classification results on the training dataset for Aug-Linear models based on MedRoBERTa.nl embeddings and logistic regression, with different numbers of n-grams and different frequency thresholds. 
%P = precision. R = recall. F1 = F1-score. AUC = Area under the receiver operating characteristic curve.
}
\label{tab:auggam-lr-thresholds}
\vspace{0.5em}
\tiny
\begin{tabular}{lrcccc}
\toprule
& Freq-threshold & \makecell{P {[}\%{]} \\ (std)} & \makecell{R {[}\%{]} \\ (std)} & \makecell{F1 {[}\%{]} \\ (std)} & \makecell{AUC {[}\%{]} \\ (std)} \\ 
\midrule
\multirow{12}{*}{\rotatebox[origin=c]{90}{Unigrams}} 
& 50 & \makecell{63.45 \\ (1.2)} & \makecell{62.41 \\ (1.3)} & \makecell{62.93 \\ (1.3)} & \makecell{73.41 \\ (0.9)} \\
& 100 & \makecell{67.85 \\ (1.2)} & \makecell{65.42 \\ (1.3)} & \makecell{66.61 \\ (1.3)} & \makecell{73.54 \\ (0.9)} \\
& 500 & \makecell{70.45 \\ (1.0)} & \makecell{68.45 \\ (1.1)} & \makecell{69.44 \\ (1.2)} & \makecell{74.12 \\ (1.2)} \\
& 1000 & \makecell{72.25 \\ (8.0)} & \makecell{69.85 \\ (1.4)} & \makecell{71.54 \\ (1.2)} & \makecell{75.54 \\ (0.9)} \\
& 5000 & \makecell{54.65 \\ (9.0)} & \makecell{55.25 \\ (1.2)} & \makecell{55.51 \\ (1.2)} & \makecell{66.41 \\ (1.1)} \\
& 10000 & \makecell{48.95 \\ (1.2)} & \makecell{44.54 \\ (0.5)} & \makecell{46.64 \\ (1.2)} & \makecell{54.21 \\ (1.0)} \\
\midrule

\multirow{12}{*}{\rotatebox[origin=c]{90}{Bigrams}} 
& 50 & \makecell{70.21 \\ (1.2)} & \makecell{68.54 \\ (1.3)} & \makecell{69.36 \\ (1.3)} & \makecell{74.21 \\ (1.0)} \\
& 100 & \makecell{71.42 \\ (1.4)} & \makecell{70.12 \\ (1.3)} & \makecell{70.76 \\ (1.3)} & \makecell{74.32 \\ (1.1)} \\
& 500 & \makecell{72.45 \\ (1.0)} & \makecell{70.87 \\ (1.0)} & \makecell{71.65 \\ (1.0)} & \makecell{74.54 \\ (12.0)} \\
& 1000 & \makecell{68.59 \\ (1.0)} & \makecell{68.97 \\ (1.2)} & \makecell{68.77 \\ (1.1)} & \makecell{74.01 \\ (0.9)} \\
& 5000 & \makecell{55.11 \\ (1.4)} & \makecell{53.00 \\ (1.0)} & \makecell{54.03 \\ (1.3)} & \makecell{67.22 \\ (1.0)} \\
& 10000 & \makecell{54.20 \\ (1.2)} & \makecell{43.25 \\ (1.3)} & \makecell{48.10 \\ (1.3)} & \makecell{54.74 \\ (1.0)} \\
\midrule

\multirow{12}{*}{\rotatebox[origin=c]{90}{Trigrams}} 
& 50 & \makecell{70.45 \\ (1.0)} & \makecell{68.45 \\ (1.0)} & \makecell{69.43 \\ (1.0)} & \makecell{75.41 \\ (1.0)} \\
& 100 & \makecell{72.45 \\ (1.2)} & \makecell{70.18 \\ (1.4)} & \makecell{71.30 \\ (1.3)} & \makecell{77.45 \\ (1.2)} \\
& 500 & \makecell{\textbf{73.45} \\ (1.0)} & \makecell{\textbf{74.54} \\ (1.3)} & \makecell{\textbf{73.99} \\ (1.3)} & \makecell{\textbf{79.01} \\ (1.1)} \\
& 1000 & \makecell{66.45 \\ (1.1)} & \makecell{65.42 \\ (1.4)} & \makecell{65.93 \\ (1.2)} & \makecell{74.56 \\ (0.8)} \\
& 5000 & \makecell{51.23 \\ (1.0)} & \makecell{50.24 \\ (1.2)} & \makecell{50.73 \\ (1.2)} & \makecell{65.41 \\ (1.3)} \\
& 10000 & \makecell{45.65 \\ (1.1)} & \makecell{44.21 \\ (1.2)} & \makecell{44.91 \\ (1.1)} & \makecell{58.41 \\ (1.1)} \\
\midrule

\multirow{12}{*}{\rotatebox[origin=c]{90}{4-Grams}} 
& 50 & \makecell{67.45 \\ (1.1)} & \makecell{69.54 \\ (1.1)} & \makecell{68.47 \\ (1.1)} & \makecell{74.12 \\ (1.0)} \\
& 100 & \makecell{68.56 \\ (1.0)} & \makecell{68.45 \\ (1.0)} & \makecell{68.50 \\ (1.0)} & \makecell{75.41 \\ (1.0)} \\
& 500 & \makecell{70.12 \\ (1.3)} & \makecell{68.79 \\ (1.0)} & \makecell{69.45 \\ (1.1)} & \makecell{76.14 \\ (0.9)} \\
& 1000 & \makecell{65.42 \\ (1.0)} & \makecell{62.32 \\ (1.2)} & \makecell{63.83 \\ (1.1)} & \makecell{70.12 \\ (1.2)} \\
& 5000 & \makecell{50.45 \\ (1.0)} & \makecell{50.24 \\ (1.1)} & \makecell{50.34 \\ (1.1)} & \makecell{64.12 \\ (1.3)} \\
& 10000 & \makecell{44.56 \\ (1.2)} & \makecell{44.21 \\ (1.1)} & \makecell{44.38 \\ (1.1)} & \makecell{59.84 \\ (1.3)} \\
\midrule

\multirow{12}{*}{\rotatebox[origin=c]{90}{5-Grams}} 
& 50 & \makecell{67.45 \\ (1.0)} & \makecell{64.58 \\ (1.2)} & \makecell{65.98 \\ (1.2)} & \makecell{71.45 \\ (1.2)} \\
& 100 & \makecell{68.94 \\ (1.3)} & \makecell{65.35 \\ (1.3)} & \makecell{67.10 \\ (1.3)} & \makecell{72.54 \\ (1.2)} \\
& 500 & \makecell{62.51 \\ (1.0)} & \makecell{64.52 \\ (1.3)} & \makecell{63.50 \\ (1.3)} & \makecell{70.14 \\ (1.3)} \\
& 1000 & \makecell{50.45 \\ (1.2)} & \makecell{63.21 \\ (1.3)} & \makecell{56.11 \\ (1.3)} & \makecell{65.48 \\ (0.9)} \\
& 5000 & \makecell{48.78 \\ (1.1)} & \makecell{50.24 \\ (1.2)} & \makecell{49.49 \\ (1.2)} & \makecell{62.11 \\ (1.2)} \\
& 10000 & \makecell{47.12 \\ (1.1)} & \makecell{49.87 \\ (1.1)} & \makecell{48.46 \\ (1.1)} & \makecell{59.74 \\ (1.1)} \\
\bottomrule
\end{tabular}
\end{table}

\begin{table}[hp]
\centering
\tiny
\caption{10-fold cross-validation classification results on the training dataset for Aug-Linear models based on MedRoBERTa.nl embeddings and explainable boosting machine, with different numbers of n-grams and different frequency thresholds.
%P = precision. R = recall. F1 = F1-score. AUC = Area under the receiver operating characteristic curve.
}
\label{tab:auggam-ebm-thresholds}
\vspace{0.5em}

\begin{tabular}{lcccccc}
\toprule
& Freq-threshold & \makecell{P [\%] \\ \footnotesize{(std)}} & \makecell{R [\%] \\ \footnotesize{(std)}} & \makecell{F1 [\%] \\ \footnotesize{(std)}} & \makecell{AUC [\%] \\ \footnotesize{(std)}} \\
\midrule

\multirow{12}{*}{\rotatebox[origin=c]{90}{Unigrams}} 
& 50    & \makecell{68.01 \\ (0.80)}  & \makecell{65.45 \\ (0.80)}  & \makecell{66.47 \\ (0.80)}  & \makecell{71.45 \\ (0.90)} \\
& 100   & \makecell{68.45 \\ (1.40)}  & \makecell{68.45 \\ (1.30)}  & \makecell{68.22 \\ (1.30)}  & \makecell{73.45 \\ (1.00)} \\
& 500   & \makecell{70.45 \\ (1.40)}  & \makecell{72.45 \\ (1.10)}  & \makecell{70.98 \\ (1.20)}  & \makecell{74.89 \\ (1.20)} \\
& 1000  & \makecell{67.24 \\ (1.20)}  & \makecell{69.85 \\ (1.10)}  & \makecell{68.40 \\ (1.20)}  & \makecell{72.41 \\ (1.00)} \\
& 5000  & \makecell{54.63 \\ (1.00)}  & \makecell{50.41 \\ (1.10)}  & \makecell{51.92 \\ (1.00)}  & \makecell{60.12 \\ (1.30)} \\
& 10000 & \makecell{52.22 \\ (1.00)}  & \makecell{49.11 \\ (1.10)}  & \makecell{50.45 \\ (1.00)}  & \makecell{50.41 \\ (1.40)} \\
\midrule

\multirow{12}{*}{\rotatebox[origin=c]{90}{Bigrams}} 
& 50    & \makecell{65.21 \\ (0.90)}  & \makecell{66.12 \\ (1.30)}  & \makecell{65.49 \\ (1.20)}  & \makecell{73.21 \\ (0.80)} \\
& 100   & \makecell{71.24 \\ (1.10)}  & \makecell{71.22 \\ (1.00)}  & \makecell{70.49 \\ (1.00)}  & \makecell{73.45 \\ (1.20)} \\
& 500   & \makecell{\textbf{72.24} \\ (1.30)}  & \makecell{71.90 \\ (1.20)}  & \makecell{71.49 \\ (1.20)}  & \makecell{73.89 \\ (1.00)} \\
& 1000  & \makecell{68.59 \\ (1.30)}  & \makecell{67.23 \\ (1.10)}  & \makecell{67.78 \\ (1.20)}  & \makecell{71.45 \\ (1.00)} \\
& 5000  & \makecell{57.24 \\ (1.20)}  & \makecell{60.12 \\ (1.40)}  & \makecell{58.46 \\ (1.30)}  & \makecell{68.54 \\ (1.10)} \\
& 10000 & \makecell{57.00 \\ (1.00)}  & \makecell{50.48 \\ (1.30)}  & \makecell{53.27 \\ (1.30)}  & \makecell{61.23 \\ (1.00)} \\
\midrule

\multirow{12}{*}{\rotatebox[origin=c]{90}{Trigrams}} 
& 50    & \makecell{68.52 \\ (1.00)}  & \makecell{72.41 \\ (1.40)}  & \makecell{69.50 \\ (1.30)}  & \makecell{77.45 \\ (1.10)} \\
& 100   & \makecell{70.04 \\ (1.00)}  & \makecell{\textbf{78.21} \\ (0.90)}  & \makecell{\textbf{73.90} \\ (1.00)}  & \makecell{\textbf{83.42} \\ (1.20)} \\
& 500   & \makecell{64.42 \\ (1.20)}  & \makecell{68.65 \\ (1.20)}  & \makecell{65.98 \\ (1.20)}  & \makecell{75.12 \\ (1.30)} \\
& 1000  & \makecell{62.87 \\ (1.00)}  & \makecell{61.10 \\ (1.40)}  & \makecell{60.98 \\ (1.20)}  & \makecell{74.12 \\ (0.70)} \\
& 5000  & \makecell{58.14 \\ (1.30)}  & \makecell{55.01 \\ (1.10)}  & \makecell{56.46 \\ (1.30)}  & \makecell{73.21 \\ (1.20)} \\
& 10000 & \makecell{45.23 \\ (1.40)}  & \makecell{48.21 \\ (1.30)}  & \makecell{46.45 \\ (1.30)}  & \makecell{70.12 \\ (1.00)} \\
\midrule

\multirow{12}{*}{\rotatebox[origin=c]{90}{4-Grams}} 
& 50    & \makecell{65.42 \\ (1.00)}  & \makecell{62.15 \\ (1.20)}  & \makecell{63.74 \\ (1.00)}  & \makecell{73.21 \\ (0.80)} \\
& 100   & \makecell{69.51 \\ (1.40)}  & \makecell{67.45 \\ (1.40)}  & \makecell{68.46 \\ (1.40)}  & \makecell{76.87 \\ (0.90)} \\
& 500   & \makecell{68.41 \\ (1.30)}  & \makecell{65.12 \\ (1.30)}  & \makecell{66.21 \\ (1.30)}  & \makecell{74.54 \\ (0.90)} \\
& 1000  & \makecell{65.89 \\ (1.40)}  & \makecell{62.54 \\ (1.00)}  & \makecell{64.17 \\ (1.20)}  & \makecell{72.12 \\ (1.00)} \\
& 5000  & \makecell{54.12 \\ (1.20)}  & \makecell{57.23 \\ (1.00)}  & \makecell{56.64 \\ (1.20)}  & \makecell{65.42 \\ (1.00)} \\
& 10000 & \makecell{55.12 \\ (1.20)}  & \makecell{53.25 \\ (1.20)}  & \makecell{54.17 \\ (1.20)}  & \makecell{60.12 \\ (12.00)} \\
\midrule

\multirow{12}{*}{\rotatebox[origin=c]{90}{5-Grams}} 
& 50    & \makecell{68.94 \\ (1.00)}  & \makecell{65.35 \\ (0.80)}  & \makecell{67.10 \\ (1.00)}  & \makecell{72.15 \\ (1.20)} \\
& 100   & \makecell{54.65 \\ (1.20)}  & \makecell{60.00 \\ (1.10)}  & \makecell{57.20 \\ (1.20)}  & \makecell{71.54 \\ (1.20)} \\
& 500   & \makecell{54.00 \\ (1.00)}  & \makecell{58.00 \\ (1.00)}  & \makecell{55.93 \\ (1.10)}  & \makecell{67.45 \\ (1.10)} \\
& 1000  & \makecell{50.45 \\ (1.40)}  & \makecell{54.00 \\ (1.30)}  & \makecell{52.16 \\ (1.40)}  & \makecell{65.12 \\ (1.10)} \\
& 5000  & \makecell{48.78 \\ (1.20)}  & \makecell{50.24 \\ (1.00)}  & \makecell{49.50 \\ (1.20)}  & \makecell{60.14 \\ (1.00)} \\
& 10000 & \makecell{47.12 \\ (1.40)}  & \makecell{49.87 \\ (1.10)}  & \makecell{48.46 \\ (1.40)}  & \makecell{54.87 \\ (1.30)} \\
\bottomrule
\end{tabular}
\end{table}

For $n$-grams embedding, we compute them using our best black-box transformer-based classifier, i.e. MedRoBERTa.nl.
These embeddings can be computed only once and stored, reusing them at inference time.

To compute explanations, we multiply each n-gram embedding by the model weight. Since a token might be part of a higher order $n$-gram, we assign it its score if this is higher than the score of the higher order $n$-gram(s). Otherwise, we assign it the score of the higher order $n$-gram with the highest score.

Hyperparameters of LR and EBM models were selected in the same way as for the LR and EBM models only on structured data (see Table \ref{tab:res-auggam-reg}). 

\begin{table}[H]
\centering
\caption{10-fold cross-validation classification results on the training dataset of the best LR Aug-GAM model with trigrams (frequency thresholds 1000, 500, 500) with different values of the regularization parameter $C = 1 / \lambda$.
%P = precision. R = recall. F1 = F1-score. AUC = Area under the receiver operating characteristic curve.
}
\label{tab:res-auggam-reg}
\vspace{0.5em}
\begin{tabular}{lcccc}
\toprule
L2 Reg (C) & \makecell{P {[}\%{]} \\ (std)} & \makecell{R {[}\%{]} \\ (std)} & \makecell{F1 {[}\%{]} \\ (std)} & \makecell{AUC {[}\%{]} \\ (std)} \\ 
\midrule
$10^{-3}$ & \makecell{72.31 \\ (1.0)} & \makecell{73.79 \\ (1.1)} & \makecell{73.00 \\ (1.0)} & \makecell{85.34 \\ (0.9)} \\
$10^{-2}$ & \makecell{72.06 \\ (1.3)} & \makecell{73.37 \\ (1.3)} & \makecell{72.67 \\ (1.3)} & \makecell{85.11 \\ (1.1)} \\
$10^{-1}$ & \makecell{71.65 \\ (1.0)} & \makecell{72.84 \\ (1.1)} & \makecell{72.56 \\ (1.0)} & \makecell{85.12 \\ (0.9)} \\
$1$ & \makecell{\textbf{72.88} \\ (1.0)} & \makecell{\textbf{73.85} \\ (1.1)} & \makecell{\textbf{73.34} \\ (1.0)} & \makecell{\textbf{85.45} \\ (0.9)} \\
$10$ & \makecell{72.20 \\ (1.1)} & \makecell{73.56 \\ (1.1)} & \makecell{72.83 \\ (1.1)} & \makecell{85.05 \\ (0.9)} \\
$10^2$ & \makecell{72.24 \\ (0.8)} & \makecell{73.53 \\ (1.0)} & \makecell{72.84 \\ (0.9)} & \makecell{85.00 \\ (0.9)} \\
$10^3$ & \makecell{72.08 \\ (1.7)} & \makecell{73.47 \\ (1.5)} & \makecell{72.72 \\ (1.5)} & \makecell{84.86 \\ (1.0)} \\
\bottomrule
\end{tabular}
\end{table}

\subsection{Explanations evaluation}
\label{app:expl-eval}
Local explanations are evaluated by computing the agreement between the ground truth explanations derived by manual annotations and the explanations produced by Aug-Linear models, LIME, and SHAP. The agreement is computed via Cohen's Kappa, Krippendorff's alpha, F1-score and Kendall's Tau. For Krippendorff's alpha, we consider ordinal labels in the following order: indication for the opposite class, no indication, strong indication for the current class, and complete giveaway for the current class.
Considering that the xAI techniques provide explanations by means of scores for tokens/$n$-grams and that the first three metrics require two sets of discrete labels to be compared, we define cutoff thresholds to convert the (normalized) scores into the 4 categories used during manual annotations. In particular, we consider complete giveaway scores $> 0.8$; strong indication scores $>0.2$ and indication for the opposite label scores $< -0.3$. Kendall's Tau instead allows for a direct comparison of numerical scores with categorical (ordinal) labels. We select this metric among those that can measure an association between a numerical and an ordinal variable since it does not require strong assumptions, such as normality of the scores, and works well even for small datasets. 

For global explanations, we compute the percentage of n-grams in the global explanations of each model that are marked as relevant by annotators.

\subsection{Additional results on explainability}
\begin{figure}
    \centering
    \includegraphics[height=.9\textheight]{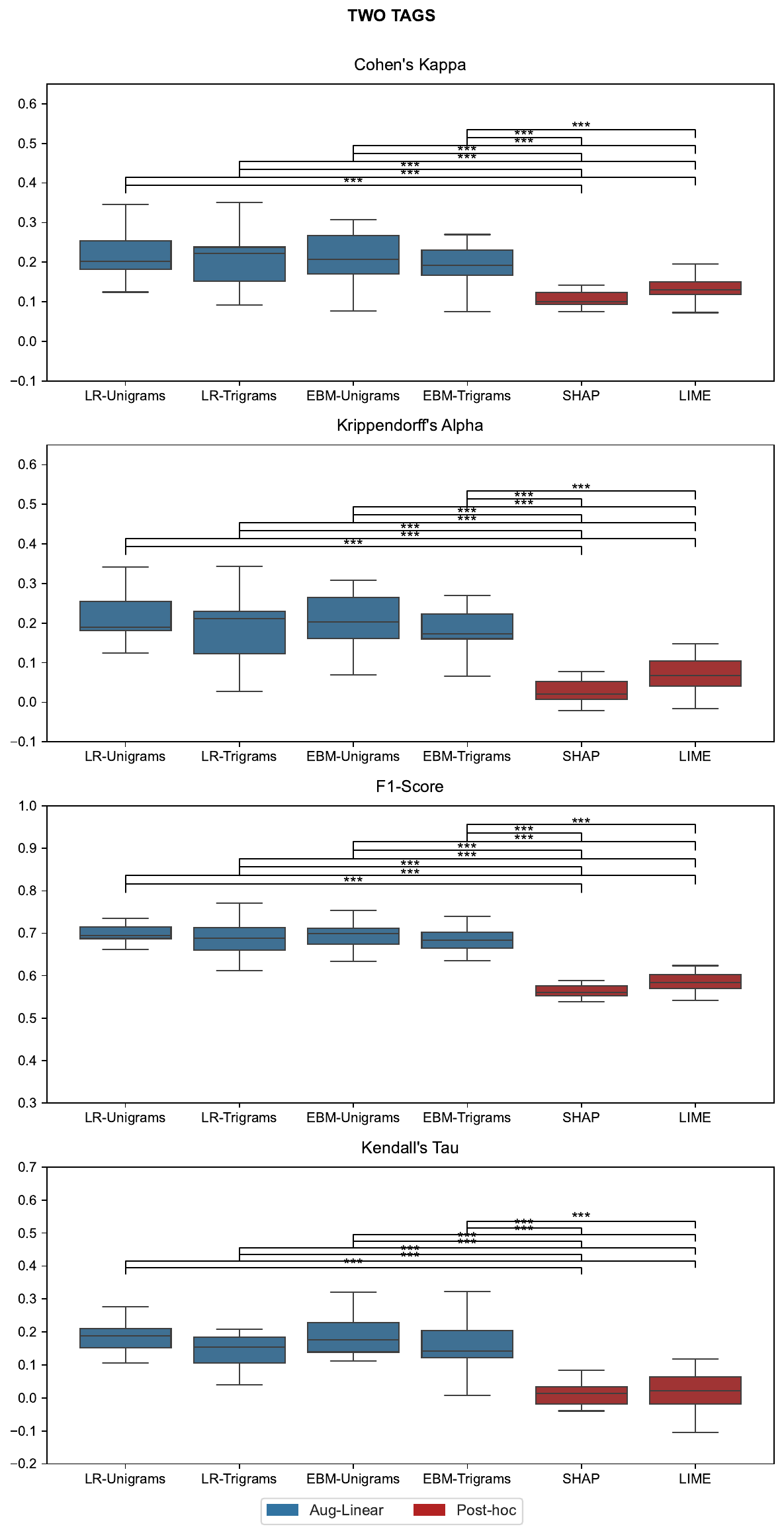}
    \caption{Results for the evaluation of local explanations, computing agreement between the different explanation methods and the annotations of
    two annotators, considering two tags: no indication and indication for the correct class. P-values of Mann-Whitney U test for differences in medians with Bonferroni correction: $*** < 0.001$}
    \label{fig:expl-two-tags}
\end{figure}

\begin{figure}
    \centering
    \includegraphics[height=.9\textheight]{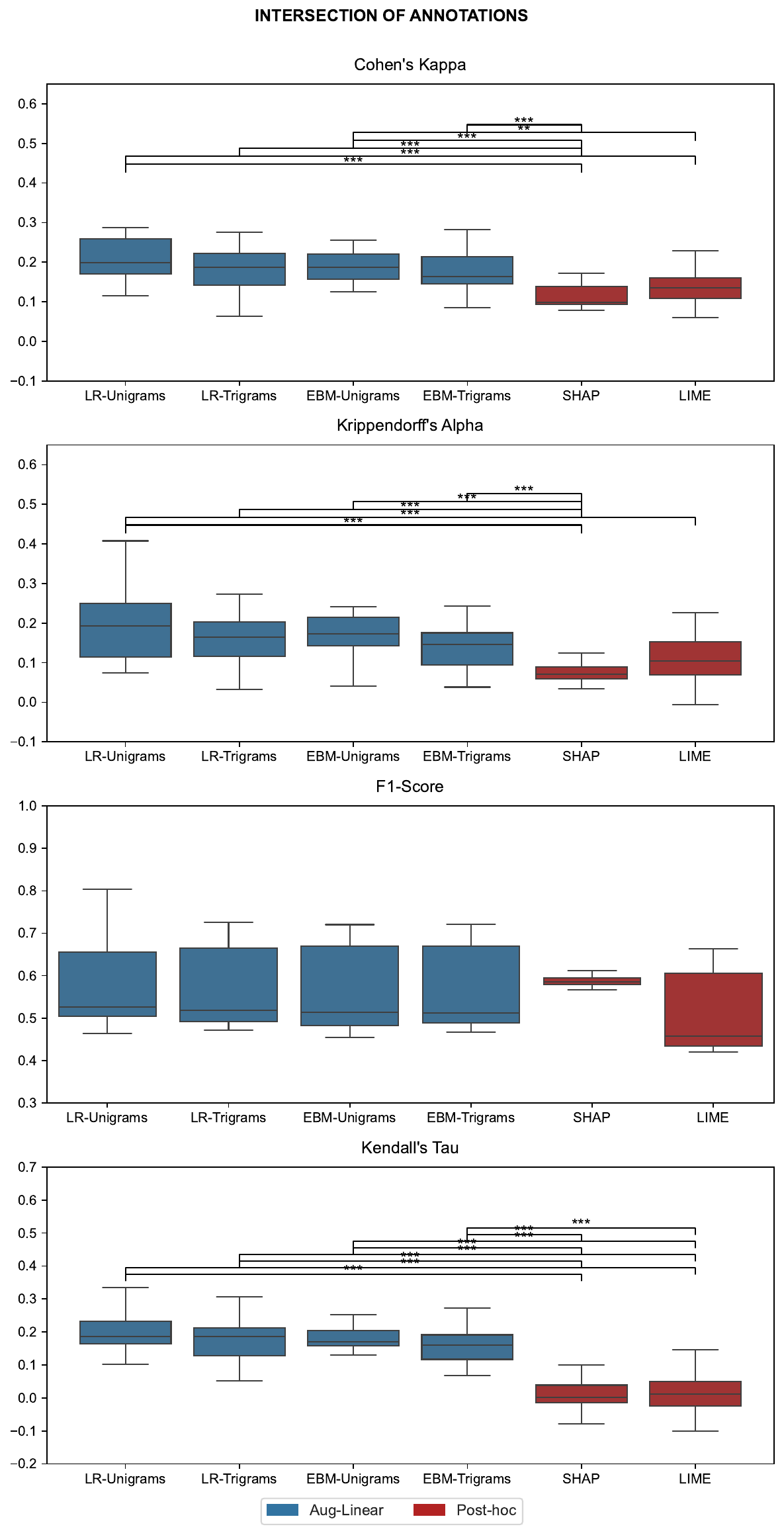}
    \caption{Results for the evaluation of local explanations, computing agreement between the different explanation methods and the intersection of the annotations of
    two annotators, considering three tags: no indication, indication for the correct class, and indication for the opposite class. P-values of Mann-Whitney U test for differences in medians with Bonferroni correction: $** < 0.01, *** < 0.001$}
    \label{fig:expl-intersect}
\end{figure}

\begin{figure}
    \centering
    \includegraphics[height=.9\textheight]{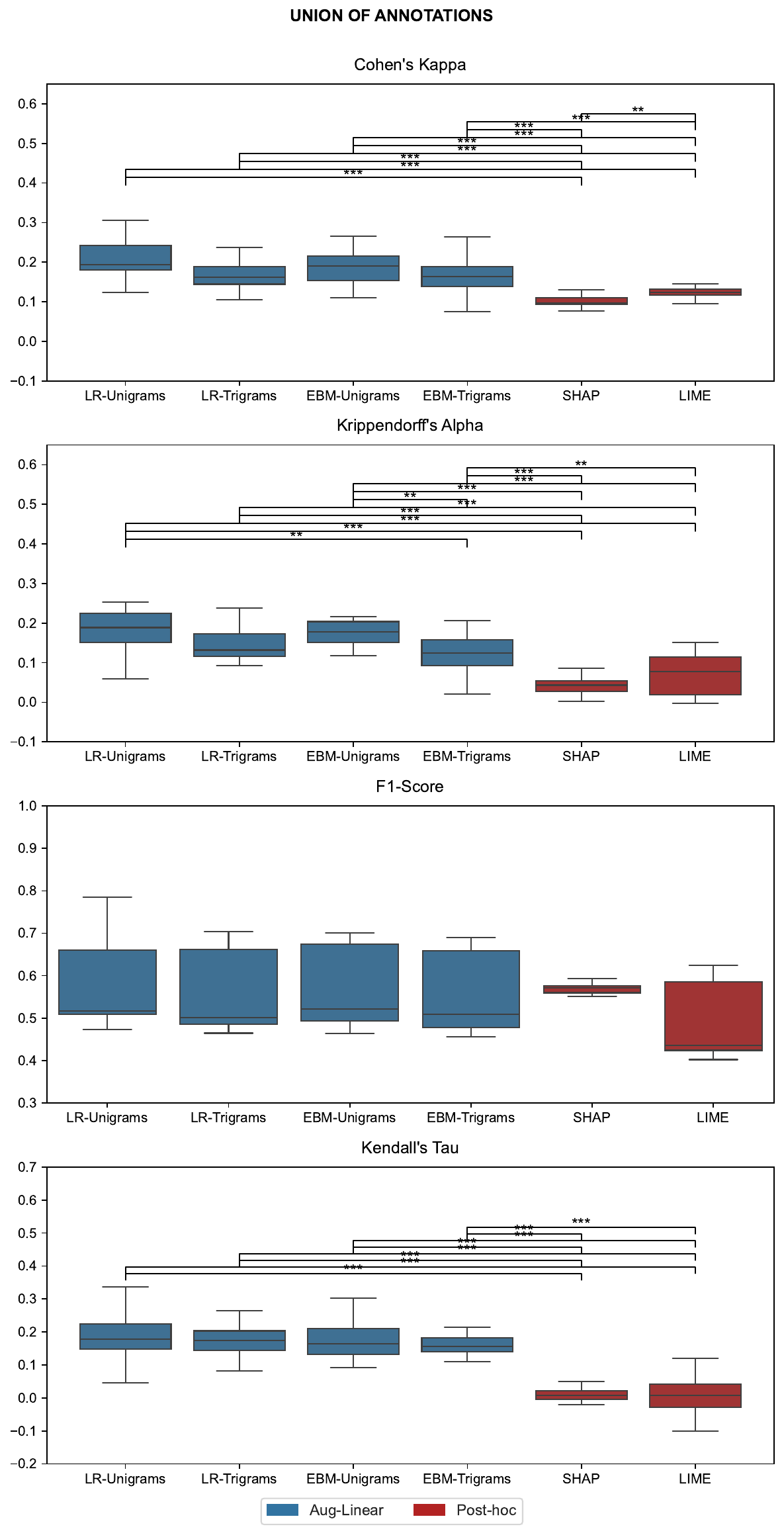}
    \caption{Results for the evaluation of local explanations, computing agreement between the different explanation methods and the union of the annotations of
    two annotators, considering three tags: no indication, indication for the correct class, and indication for the opposite class. P-values of Mann-Whitney U test for differences in medians with Bonferroni correction: $** < 0.01, *** < 0.001$}
    \label{fig:expl-union}
\end{figure}

Figures \ref{fig:expl-two-tags}, \ref{fig:expl-intersect} and \ref{fig:expl-union} report additional results on the local explanation comparison, considering only two tags (indication for the current class and no indication) and considering three tags with the intersection and the union of the annotations of the two annotators, instead of our manual merging.

\begin{table*}
\centering
\caption{Global explanations as the 15 most relevant unigrams for HFrEF (upper part) and HFpEF (lower part) for Aug-Linear models with unigrams. Green backgrounds are those assessed as clinically relevant.}
\label{tab:global-expl-auglinear-uni}
\footnotesize
\begin{tabular}{lllll}
\toprule
 \multicolumn{1}{c}{\multirow{2}{*}{\#}} &
  \multicolumn{2}{c}{\textbf{Aug-Linear LR UNI (HFrEF)}} &
  \multicolumn{2}{c}{\textbf{Aug-Linear EBM UNI (HFrEF)}} \\
\multicolumn{1}{c}{} &
  \multicolumn{1}{c}{Unigram {[}NL{]}} &
  \multicolumn{1}{c}{Unigram {[}ENG{]}} &
  \multicolumn{1}{c}{Unigram {[}NL{]}} &
  \multicolumn{1}{c}{Unigram {[}ENG{]}} \\ \midrule
1 &
  \cellcolor[HTML]{8ED973}matigslechte &
  \cellcolor[HTML]{8ED973}moderately bad &
  \multicolumn{1}{l}{\cellcolor[HTML]{8ED973}bumetanide} &
  \cellcolor[HTML]{8ED973}bumetanide \\
2 &
  \cellcolor[HTML]{8ED973}gepaced &
  \cellcolor[HTML]{8ED973}paced &
  \multicolumn{1}{l}{ezetrol} &
  ezetrol \\
3 &
  hoofdingang &
  main &
  \multicolumn{1}{l}{hoev} &
  howv \\
4 &
  medicamenteuze &
  medicated &
  \multicolumn{1}{l}{{opgeloste}} &
  dissolved \\
5 &
  richtingen &
  directions &
  \multicolumn{1}{l}{{fluticason}} &
  fluticasone \\
6 &
  semiarts &
  semi doctor &
  \multicolumn{1}{l}{druppels} &
  drops \\
7 &
  {meermaals} &
  {multiple} &
  \multicolumn{1}{l}{\cellcolor[HTML]{8ED973}enalapril} &
  \cellcolor[HTML]{8ED973}enalapril \\
8 &
  inkomend &
  incoming &
  \multicolumn{1}{l}{dr} &
  dr \\
9 &
  {gepoogd} &
 {attempted} &
  \multicolumn{1}{l}{mylan} &
  mylan \\
10 &
  ingang &
  entry &
  \multicolumn{1}{l}{werd} &
  was \\
11 &
  \cellcolor[HTML]{8ED973}slecht &
  \cellcolor[HTML]{8ED973}poorly &
  \multicolumn{1}{l}{natriumfosfaten} &
  sodium phosphates \\
12 &
  voortgeleid &
  routed &
  \multicolumn{1}{l}{\cellcolor[HTML]{8ED973}thuismedicatie} &
  \cellcolor[HTML]{8ED973}home medication \\
13 &
  ohm &
  ohm &
  \multicolumn{1}{l}{zon} &
  sun \\
14 &
  vpk &
  vpk &
  \multicolumn{1}{l}{vlgs} &
  vlgs \\
15 &
  \cellcolor[HTML]{8ED973}{slechte} &
  \cellcolor[HTML]{8ED973}{bad} &
  \multicolumn{1}{l}{pleurale
} &
  {pleural} \\ \midrule
 \multicolumn{1}{c}{\multirow{2}{*}{\#}} &
  \multicolumn{2}{c}{\textbf{Aug-Linear LR UNI (HFpEF)}} &
    \multicolumn{2}{c}{\textbf{Aug-Linear EBM UNI (HFpEF)}} \\
\multicolumn{1}{c}{} &
  \multicolumn{1}{c}{Unigram {[}NL{]}} &
  \multicolumn{1}{c}{Unigram {[}ENG{]}} &
  \multicolumn{1}{c}{Unigram {[}NL{]}} &
  \multicolumn{1}{c}{Unigram {[}ENG{]}} \\ \midrule
1 &
  totale &
  total &
  \multicolumn{1}{l}{transcatheter} &
  transcatheter \\
2 &
  \cellcolor[HTML]{8ED973}respiratoir &
  \cellcolor[HTML]{8ED973}respiratory &
  \multicolumn{1}{l}{\cellcolor[HTML]{8ED973}vermoeidheidsklachten} &
  \cellcolor[HTML]{8ED973}fatigue symptoms \\
3 &
  \cellcolor[HTML]{8ED973}behoud &
  \cellcolor[HTML]{8ED973}preserved &
  \multicolumn{1}{l}{geheugenklachten} &
  memory complaints \\
4 &
  \cellcolor[HTML]{8ED973}ejectie &
  \cellcolor[HTML]{8ED973}ejection &
  \multicolumn{1}{l}{diagnostiek} &
  diagnostics \\
5 &
  retrograad &
  retrograde &
  \multicolumn{1}{l}{\cellcolor[HTML]{8ED973}ejectiefractie} &
  \cellcolor[HTML]{8ED973}ejection fraction \\
6 &
  voorbehoud &
  caveat &
  \multicolumn{1}{l}{reactievermogen} &
  responsiveness \\
7 &
  intervalanamnese &
  interval anamnesis &
  \multicolumn{1}{l}{oorzaak} &
  cause \\
8 &
  wortel &
  root &
  \multicolumn{1}{l}{geworden} &
  become \\
9 &
  relevant &
  relevant &
  \multicolumn{1}{l}{geheugenproblemen} &
  memory problems \\
10 &
  eosinofilie &
  eosinophilia &
  \multicolumn{1}{l}{geringspap} &
  geringspap \\
11 &
  {valneiging} &
  {fall tendency} &
  \multicolumn{1}{l}{plaatsing} &
  placement \\
12 &
  instabiele &
  unstable &
  \multicolumn{1}{l}{afwijkingengeen} &
  abnormalities no \\
13 &
  transthoracale &
  transthoracic &
  \multicolumn{1}{l}{plaatselijke} &
  local \\
14 &
  \cellcolor[HTML]{8ED973}ventrikelvolgrespons &
  \cellcolor[HTML]{8ED973}\begin{tabular}[c]{@{}l@{}}ventricular \\ tracking response\end{tabular} &
  \multicolumn{1}{l}{wetenschappelijk} &
  scientific \\
15 &
  beeldkwaliteit &
  image quality &
  \multicolumn{1}{l}{depressieve} &
  depressive \\ \hline
\end{tabular}
\end{table*}

\begin{table*}
\centering
\caption{Global explanations as the 15 most relevant n-grams for HFrEF (upper part) and HFpEF (lower part) for SHAP and LIME. Green backgrounds are those assessed as clinically relevant.}
\label{tab:global-expl-shap-lime}
\footnotesize
\begin{tabular}{lllll}
\toprule 
\multirow{2}{*}{\#}
 &
  \multicolumn{2}{c}{\textbf{LIME (HFrEF)}} &
  \multicolumn{2}{c}{\textbf{SHAP (HFrEF)}} \\
\multicolumn{1}{c}{} &
  \multicolumn{1}{c}{Unigram {[}NL{]}} &
  \multicolumn{1}{c}{Unigram {[}ENG{]}} &
  \multicolumn{1}{c}{Unigram {[}NL{]}} &
  \multicolumn{1}{c}{Unigram {[}ENG{]}} \\ \midrule
\multicolumn{1}{l}{1} &
  opname &
  recording &
  \multicolumn{1}{l}{intensivist} &
  \multicolumn{1}{l}{intensivist} \\
2 &
  lca &
  lca &
  \multicolumn{1}{l}{hartcatheterisatie} &
  \multicolumn{1}{l}{cardiac catheterisation} \\
3 &
  ntie &
  ntion &
  \multicolumn{1}{l}{anteroseptaal} &
  \multicolumn{1}{l}{anteroseptal} \\
4 &
  hartfalen &
  heart failure &
  \multicolumn{1}{l}{septaal} &
  \multicolumn{1}{l}{septal} \\
5 &
  number &
  number &
  \multicolumn{1}{l}{engels} &
  \multicolumn{1}{l}{english} \\
6 &
  uitgebreid &
  extensive &
  \multicolumn{1}{l}{spreekt} &
  \multicolumn{1}{l}{speaks} \\
7 &
  ernstig &
  severe &
  \multicolumn{1}{l}{gemobiliseerd} &
  \multicolumn{1}{l}{mobilised} \\
8 &
  mid &
  mid &
  \multicolumn{1}{l}{pocket} &
  \multicolumn{1}{l}{pocket} \\
9 &
  \cellcolor[HTML]{8ED973}matige &
  \cellcolor[HTML]{8ED973}moderate &
  \multicolumn{1}{l}{\cellcolor[HTML]{8ED973}{slechte}} &
  \multicolumn{1}{l}{\cellcolor[HTML]{8ED973}{bad}} \\
10 &
  urgentie &
  urgency &
  \multicolumn{1}{l}{mim} &
  \multicolumn{1}{l}{mim} \\
11 &
  \cellcolor[HTML]{8ED973}ischemische &
  \cellcolor[HTML]{8ED973}ischemic &
  \multicolumn{1}{l}{dochter} &
  \multicolumn{1}{l}{daughter} \\
12 &
  mg &
  mg &
  \multicolumn{1}{l}{aangetroffen} &
  \multicolumn{1}{l}{found} \\
13 &
  nte &
  nte &
  \multicolumn{1}{l}{mede} &
  \multicolumn{1}{l}{co} \\
14 &
  links &
  left &
  \multicolumn{1}{l}{namens} &
  \multicolumn{1}{l}{on behalf of} \\
15 &
  geworden &
  become &
  \multicolumn{1}{l}{emboliebron} &
  \multicolumn{1}{l}{embolic source} \\ \midrule
 \multirow{2}{*}{\#} &
  \multicolumn{2}{c}{\textbf{LIME (HFrEF)}} &
  \multicolumn{2}{c}{\textbf{SHAP (HFrEF)}} \\
\multicolumn{1}{c}{} &
  \multicolumn{1}{c}{Unigram {[}NL{]}} &
  \multicolumn{1}{c}{Unigram {[}ENG{]}} &
  \multicolumn{1}{c}{Unigram {[}NL{]}} &
  \multicolumn{1}{c}{Unigram {[}ENG{]}} \\ \midrule
\multicolumn{1}{l}{1} &
  ondertekend &
  signed &
  pagina &
  page \\
2 &
  icva &
  icva &
  beschreven &
  described \\
3 &
  \cellcolor[HTML]{8ED973}afib &
  \cellcolor[HTML]{8ED973}atrial fibrillation &
  willen &
  want \\
4 &
  voorstel &
  proposal &
  na &
  after \\
5 &
  medicatie &
  medication &
  recent &
  recent \\
6 &
  hartas &
  cardiac axis &
  co &
  co \\
7 &
  cordis &
  cordis &
  coloscopie &
  colonoscopy \\
8 &
  oraal &
  oral &
  coecum &
  coecum \\
9 &
  \cellcolor[HTML]{8ED973}dapaglifozine &
  \cellcolor[HTML]{8ED973}dapaglifozine &
  sigmoid &
  sigmoid \\
10 &
  chirurgie &
  surgery &
  diverticulose &
  diverticulosis \\
11 &
  waarbij &
  where &
  geb &
  geb \\
12 &
  rvhgeen &
  rvh no &
  verwijderd &
  removed \\
13 &
  \cellcolor[HTML]{8ED973}ef &
  \cellcolor[HTML]{8ED973}ef &
  koude &
  cold \\
14 &
  septum &
  septum &
  onderwerp &
  subject \\
15 &
  lcx &
  lcx &
  hartfalen &
  heart failure \\ \bottomrule
\end{tabular}
\end{table*}

Tables \ref{tab:global-expl-auglinear-uni} and \ref{tab:global-expl-shap-lime} report the global explanations produced by the Aug-linear models with unigrams and by the post-hoc explanations methods, respectively. 

%\clearpage

\section{Appendix E - Manual annotations for explanations}
\setcounter{table}{0}    
\label{app:E}
\subsection{Local explanations}
To derive consistent and reliable annotations to be used for the evaluation and comparison of the different local explanation methods, two clinicians (M.V.C and D.K.) were asked to annotate 20 documents randomly selected in the dataset. They were requested to highlight words or groups of consecutive words corresponding to:
\begin{itemize}
    \item a strong indication for the class
    \item a complete giveaway for the class
    \item an indication for the opposite class
\end{itemize}
These annotations were collected with Microsoft Word, using three different colors to highlight words/groups of consecutive words corresponding to these three categories. 

We initially defined a set of guidelines for these annotations. After the annotation of the first 4 documents was completed, these annotations were revised, addressing issues and inconsistencies. A single annotated version of the documents was derived, to be used for evaluation, and guidelines were expanded after discussion with annotators. This process was iteratively repeated with a subsequent batch of 4 documents, followed by other 6 documents and by the last 6 documents. 
The strong indication and the complete giveaway categories were subsequently merged after a discussion with clinicians that highlighted the complexity of discrimination between the two categories.

Table~\ref{tab:ia-agreement-3} reports the inter-annotator agreement evolution along the annotation rounds. Below we report the last version of the guidelines.

\begin{table}[]
\centering
\caption{Evolution of inter-annotator agreement metrics along the annotation rounds. Metrics are computed with lenient matching, considering three tags (no indication, indication for the correct class, indication for the opposite class).}
\label{tab:ia-agreement-3}
\begin{tabular}{crrr}
\toprule
Round \# & Cohen's Kappa & Krippendorff's Alpha & F1-Score \\ \midrule
1        & 0.2057        & 0.1789               & 0.4056   \\
2        & 0.2704        & 0.2028               & 0.4121  \\ 
3        & 0.3562        & 0.4014               & 0.5106  \\
4        & 0.3843        & 0.4215               & 0.5184  \\ 
\bottomrule
\end{tabular}
\end{table}

\subsubsection*{ANNOTATION GUIDELINES FOR LOCAL EXPLANATIONS}
\paragraph{Annotation Examples}
There are several instructions in the guideline, and each is followed by one or more examples. Examples focus on the subject of the instruction. If another relevant concept in the sentence is not annotated (highlighted), it simply means that it is not the focus of the example and not that it should not be annotated in practice.

\paragraph{Annotation Procedure}
The cases will be provided in a Word document with the label stated in the header of the document. Given the patient label (HFpEF or HFrEF), highlight the terms that:

\begin{itemize}
    \item Completely give away the label (\hldarkgreen{dark green}).
    \item Give a strong suggestion of the label (\hllightgreen{light green}).
    \item Contradict the label (\hlred{red}).
\end{itemize}

A mention in the clinical notes \textit{that completely gives away the label} should be used when it makes it certain to you that this patient has the label. For example, for a patient with the label HFpEF, by reading the mention of diastolic heart failure in the clinical note, you are certain that the patient has HFpEF.

Punctuation characters such as commas (,), full stops (.), parentheses (()), and hyphens (-), or forward slashes (/) that are not part of the mention should not be included. Spaces and punctuation may be included if they are part of the mention. For example, the multi-word concept \textit{acute on chronic nierinsufficientie} includes spaces in its span, and the concept \textit{iv-contrast} contains a hyphen.

\paragraph{Categories and Mentions to Annotate}

\paragraph{Mentions that Can Suggest the Type of HF}
\begin{itemize}
    \item \textbf{Text:} "Reden van opname \hllightgreen{Linkszijdige decompensatio cordis}"
    \\ \textbf{Explanation:} This indicates an admission for acute deterioration of cardiac function, which is more likely to indicate HFrEF.

    \item \textbf{Text:} "bij het instellen op \hllightgreen{hartfalen medicatie} gedurende \hllightgreen{3 maanden} zodat er een beoordeling mogelijk is om een \hllightgreen{ICD indicatie} te stellen"
    \\ \textbf{Explanation:} In HFrEF it is common to put patients on HF-specific medications and check if patients recover or not after 3 months, after which they can receive an ICD.

    \item \textbf{Text:} "2015 Ernstige aortaklepstenose met ernstige calcificatie, \hllightgreen{goede linker ventrikelfunctie}. CAG: Natief drievatslijden, functie grafts goed met uitzondering Ao-D2 (afgesloten). D2 wordt gevuld via Ao-D1-MO1  2016 (2) Ongecompliceerde TAVI. Geringe paravalvulaire lekkage. Gering tot matige mitralisklepinsufficientie."
    \\ \textbf{Explanation:} In HFpEF, LV ejection fraction is preserved. Even though this is in the history, it already gives an indication that this patient may have HFpEF instead of HFrEF as LV function is explicitly stated.

    \item \textbf{Text:} "Op echo \hldarkgreen{goede LV functie en een mitralisklep insufficientie gr. II}, \hllightgreen{TI. gr II}. Aanvullende MPS toonde EF 73\% en \hllightgreen{dubieus minimale ischemie septaal}."
    \\ \textbf{Explanation:} The combination of mentioning a good LV function combined with MI is very suggestive of HFpEF and thus these factors should be marked together as one annotation.

    \item \textbf{Text:} "Conclusie: Geen aperte aanwijzingen voor decompensatio cordis. Asymmetrisch oedeem mgl. door veneuze insufficientie na venectomie. Dyspnoe mogelijk op basis van \hldarkgreen{diastolische dysfunctie}."
    \\ \textbf{Explanation:} The mention of diastolic dysfunction in relation to the symptoms is a clear indication that the patient has HFpEF.
\end{itemize}

\paragraph{Known Underlying Causes/Comorbidities Related to HFrEF/HFpEF}
\begin{itemize}
    \item \textbf{Text:} "Rechts- en linkszijdig hartfalen bij \hllightgreen{dilaterende cardiomyopathie} de novo"
    \\ \textbf{Explanation:} This indicates the underlying cause that is unambiguous for HFrEF.

    \item \textbf{Text:} "Het betreft een [LEEFTIJD-1]-jarige patiente, bekend met hypertensie, diabetes, hypercholesterolemie, chronische nierinsufficientie, proximale myopathie waarschijnlijk op basis van SCN4A mutatie, \hllightgreen{persirend AF} en \hldarkgreen{diastolisch hartfalen}."
    \\ \textbf{Explanation:} Comorbidities related to HFpEF and the explicit mention of diastolic heart failure.
\end{itemize}

\paragraph{Explicit LVEF Mentions}
\begin{itemize}
    \item \textbf{Text:} "\hldarkgreen{LVEF 19\%}."
    \\ \textbf{Explanation:} Explicit mentions of LVEF $< 40\%$ are a clear indication of HFrEF.

    \item \textbf{Text:} "Echocardiografisch werd er mogelijk een low flow, low gradient severe AS (bij \hldarkgreen{slechte LV functie}, ernstige MI en TI)."
    \\ \textbf{Explanation:} Explicit mentions of poor LVEF giving the clear indication of HFrEF.

    \item \textbf{Text:} "Linkerventrikel: Ernstige concentrische \hllightgreen{linker ventrikel hypertrofie} met \hldarkgreen{goede systolische functie}, klein systolisch volume, \hldarkgreen{diastolische dysfunctie graad II}."
    \\ \textbf{Explanation:} Explicit mentions of good systolic function and diastolic dysfunction grading giving the clear indication of HFpEF. Additionally, left ventricular hypertrophy is mentioned, which is a common underlying cause of HFpEF.
\end{itemize}

\paragraph{Medications}
\begin{itemize}
    \item \textbf{Text:} "\hllightgreen{metoPROLOL} tartraat 50 mg tablet"
    \\ \textbf{Explanation:} Medications that are common general heart failure therapy, but are more frequent in HFrEF.

    \item \textbf{Text:} "Patient kreeg \hldarkgreen{furosemide intraveneus} waarop goed resultaat."
    \\ \textbf{Explanation:} This indicates medication treatment only given when patients with HFrEF are hospitalized for acute HF decompensation, closely related to HFrEF.

    \item \textbf{Text:} "Thuismedicatie (voor zover bij mij bekend) - \hllightgreen{bumetanide} 1 mg tablet, 1 mg, oraal, 1dd ZN - \hllightgreen{dapagliflozine} 10 mg TABLET tablet, 10 mg, oraal, 1dd - gliclazide 80 mg tablet MGA, 80 mg, oraal, 1dd - insuline glargine (LANTUS) 100 IE/ml penfillr, Injecteer onder de huid - metformine (METFORMINE) 1000 mg tablet, 1.000 mg, oraal, 3dd - \hllightgreen{metoPROLOL} SUCCINaat 50 mg tablet MGA, 75 mg, oraal, 1dd - omeprazol 20 mg capsule MSR, 20 mg, oraal, 1dd ZN - psylliumvezels 3,25 g granulaat, 1 sachet, oraal, 1dd ZN - \hllightgreen{sacubitril/valsartan} 24/26 mg (\hllightgreen{ENTRESTO}) tablet, 1 tablet, 2dd - \hllightgreen{spironolacton} 25 mg tablet, 25 mg, oraal, 1dd"
    \\ \textbf{Explanation:} In cases where the medication dosage does not matter, but having (a combination) of drugs indicates the type of heart failure, then only the name of the drug can be marked. If the dosage is relevant for specific HF types, then also the dosage should be included.
\end{itemize}

\paragraph{Outcomes of Clinical Tests Possibly Related to the Type of HF}
\begin{itemize}
    \item \textbf{Text:} "genetisch met 2 unclassified variants in \hllightgreen{TTN-gen}."
    \\ \textbf{Explanation:} Generally, genetic testing is not done in HFpEF, providing a clear indication for HFrEF in this case. Additionally, TTN is a gene associated with HFrEF.

    \item \textbf{Text:} "[PERSOON-2] LV-functie. \hldarkgreen{Diastolische dysfunctie graad II}. Abnormale septumbeweging (bij LBTB)."
    \\ \textbf{Explanation:} To confirm HFpEF, echocardiography is performed where the degree of diastolic dysfunction is measured. This is reported in the letter.
\end{itemize}

\paragraph{Signs and Symptoms Related to the Specific HF Type (HFrEF/HFpEF)}
\begin{itemize}
    \item \textbf{Text:} "Anamnese  Sinds 3 weken hllightgreen{progressief dyspnoeisch}. \hllightgreen{Dikke onderbenen} bemerkt, maar heeft \hllightgreen{steunkousen} bij vermeende veneuze insufficientie. \hllightgreen{Boller wordende buik}. \hllightgreen{Orthopneu+}. \hllightgreen{Nycturie+}."
    \\ \textbf{Explanation:} The combination of different signs and symptoms clearly indicate HFrEF.

    \item \textbf{Text:} "2. \hllightgreen{NSVTs} gedurende de opname wv start \hllightgreen{amiodarone}."
    \\ \textbf{Explanation:} This indicates medication treatment only given when patients with HFrEF and having the specific rhythm disorder (NSTVs).

    \item \textbf{Text:} "\hllightgreen{NATRIUM [INSTELLING-1] 127 (L)}   KALIUM [INSTELLING-1] [DATUM-6] (H)   \hllightgreen{KREATININE [INSTELLING-1] 217 (PH)}   \hllightgreen{EGFR (MDRD) [INSTELLING-1] 18}   UREUM [INSTELLING-1] [DATUM-7]"
    \\ \textbf{Explanation:} Measurements with values indicating the type or severity of HF. These can be tricky due to masked information, but stated values aligning with the diagnosis can be considered.

    \item \textbf{Text:} "Pulmones: normaal ademgeruis, rechts mild \hllightgreen{crepiteren}"
    \\ \textbf{Explanation:} Sign crepitation specifically indicates fluid in the lungs. The severity or side is less relevant.

    \item \textbf{Text:} "Geen pijn op de borst, geen misselijkheid, zweten of braken, het lijkt niet op zijn hartinfarct van eerder. \hlred{Geen dyspnoe}. Geen neurologische uitval."
    \\ \textbf{Explanation:} No dyspnea indicates no left-sided heart failure. It can still be right-sided, but in the case of LVHF, there should be pulmonary edema.

    \item \textbf{Text:} "Algemeen: geen koorts gehad, \hllightgreen{stabiel gewicht}, vroeger [LEEFTIJD-1] jaar in Indonesie gewoond tijdens de oorlog, hier vaak ziek geweest."
    \\ \textbf{Explanation:} Mention of stable weight indicates that the doctor looks for weight changes, which is specific for HFpEF assessment.
\end{itemize}

\paragraph{Categories and Mentions Not to Annotate}
In this section, we will show examples of information that should not be annotated as they are too general and not specific or relevant for the type of HF. These examples highlight the nuances and complexities to ensure consistency within the annotation process.

\begin{itemize}
    \item \textbf{Text:} "Datum  Ons kenmerk  Pagina       [DATUM-1]      1 van 4              [PERSOON-3], geb. [DATUM-2], gesl. vrouw, patnr. [PATIENTNUMMER-1],"
    \\ \textbf{Explanation:} These are general statements on patient characteristics, which are not specific for HF patients only.

    \item \textbf{Text:} "Bovenstaande patient lag opgenomen van [DATUM-1] tot [DATUM-1] op de afdeling cardiologie van het [INSTELLING-1]"
    \\ \textbf{Explanation:} This is a general statement which can also be true in the case of other cardiac problems, not specific for only HF patients.

    \item \textbf{Text:} "Met collegiale hoogachting, [PERSOON-1], coassistent   [PERSOON-2], arts-assistent cardiologie[PERSOON-3], cardioloog Cc: Geen ontvangers"
    \\ \textbf{Explanation:} These are general endings of the cardiology letters, but also for other diseases such letters are generated. It is not specific only for HF patients.

    \item \textbf{Text:} "NATRIUM [INSTELLING-1] 127 (L)   KALIUM [INSTELLING-1] [DATUM-6] (H)   KREATININE [INSTELLING-1] 217 (PH)   EGFR (MDRD) [INSTELLING-1] 18   UREUM [INSTELLING-1] [DATUM-7]"
    \\ \textbf{Explanation:} Similarly to the information above, the statement on general laboratory measurements without the value, being not specifically only measured in HF patients should not be marked.

    \item \textbf{Text:} "Laboratorium: [DATUM-1] 19:28   CRP [INSTELLING-1] [DATUM-2] (H)   HEMOGLOBINE [INSTELLING-1] [DATUM-3]   HEMATOCRIET [INSTELLING-1] 0.37   TROMBOCYTEN [INSTELLING-1] 165"
    \\ \textbf{Explanation:} Even though hemoglobin levels can tell something about the severity of HF, in this case, it is not marked as no value is available.

    \item \textbf{Text:} "Lichamelijk onderzoek  Niet zieke, heldere vrouw. Dyspnoisch bij spreken  Bloeddruk 153/86 mmHg, pols 80/min, temperatuur 37,5 C, SpO2 95\% bij kamerlucht."
    \\ \textbf{Explanation:} These parameters are not specific for heart failure and its type, so measurements like these should not be annotated.

    \item \textbf{Text:} "Geachte collega,   [PERSOON-5] was opgenomen op [INSTELLING-1] INTENSIVE CARE VOLWASSENEN"
    \\ \textbf{Explanation:} No need to mark the admission department as this is not specific for heart failure.

    \item \textbf{Text:} "Meet nu bloeddrukken van 1[DATUM-2] mmHg/75 mmHg."
    \\ \textbf{Explanation:} Measurements without numbers should not be annotated as they do not indicate the label.
\end{itemize}

\subsection{Global explanations}
To evaluate global explanations, the same clinicians who annotated the documents for local explanations were asked to review the global explanations produced by each model. For each of the two classes, the top 15 relevant n-grams per model were selected, yielding a total of 90 n-grams. These n-grams were presented to the annotators in random order without indicating which model produced each one. Annotators were asked to label each n-gram as \textit{relevant} or \textit{not relevant} to its associated class. 
The two resulting sets of annotations were then manually reviewed and adjudicated by all authors to produce a final version to be used for the evaluation.
In three cases, $n$-grams that had not been marked as relevant by either annotator were marked as relevant following this review. 
%behoud, ejectie, comnclusie normaal aspect
Inter-annotator agreement, measured using the same metrics as for local explanations, resulted in a Cohen’s Kappa of 0.6427, an F1-score of 0.8204, and a Krippendorff’s Alpha of 0.6428.

\end{appendices}

\end{document}